\pdfoutput=1
\documentclass{article}

\usepackage{microtype}
\usepackage{graphicx}
\usepackage{subfigure}
\usepackage{booktabs} 

\usepackage{hyperref}
\usepackage{amsmath,amssymb,amsfonts}
\usepackage{graphicx}
\usepackage[svgnames]{xcolor}
\usepackage{mdframed} 
\usepackage{listings} 
\usepackage{slashed}
\usepackage{subcaption} 

\usepackage{amsmath}  
\DeclareMathOperator{\erfc}{erfc}

\usepackage[accepted]{icml2025}

\usepackage{amsmath}
\usepackage{amssymb}
\usepackage{mathtools}
\usepackage{amsthm}
\usepackage{xcolor}
\usepackage[capitalize,noabbrev]{cleveref}

\theoremstyle{plain}

\theoremstyle{definition}

\theoremstyle{remark}

\usepackage{listings}

\icmltitlerunning{}

\lstdefinestyle{mypython}{
    language=Python,
    basicstyle=\ttfamily\scriptsize,
    keywordstyle=\color{blue},
    stringstyle=\color{red},
    commentstyle=\color{ForestGreen},
    frame=single,      framerule=0.4pt,
    numbers=none,      
    breaklines=true,   breakatwhitespace=true,
    numbersep=4pt,     tabsize=2,
    captionpos=b,
    escapeinside={(*@}{@*)}
}
\lstdefinestyle{sympyout}{
    basicstyle=\ttfamily\footnotesize,  
    frame=none,
    breaklines=true,                 
    breakatwhitespace=false,         
    columns=fullflexible, 
    keepspaces=true       
}
\mdfdefinestyle{solutionbox}{
    linecolor=blue,
    linewidth=1pt,
    frametitlerule=true,
    frametitlebackgroundcolor=blue!10,
    innertopmargin=10pt,  innerbottommargin=10pt,
    innerleftmargin=8pt,  innerrightmargin=8pt,
    leftmargin=5pt,       rightmargin=5pt,
    roundcorner=4pt,
    skipabove=\baselineskip, skipbelow=\baselineskip,
}

\mdfdefinestyle{sympybox}{
    linecolor=ForestGreen,
    linewidth=1pt,
    frametitlerule=true,
    frametitlebackgroundcolor=ForestGreen!10,
    innertopmargin=10pt,  innerbottommargin=10pt,
    innerleftmargin=8pt,  innerrightmargin=8pt,
    leftmargin=5pt,       rightmargin=5pt,
    roundcorner=4pt,
    skipabove=\baselineskip, skipbelow=\baselineskip,
}

\mdfdefinestyle{agentfeedbackbox}{
    linecolor=DarkOrchid,
    linewidth=1pt,
    frametitlerule=true,
    frametitlebackgroundcolor=DarkOrchid!10,
    innertopmargin=10pt,  innerbottommargin=10pt,
    innerleftmargin=8pt,  innerrightmargin=8pt,
    leftmargin=5pt,       rightmargin=5pt,
    roundcorner=4pt,
    skipabove=\baselineskip, skipbelow=\baselineskip,
}

\begin{document}

\twocolumn[

\icmltitle{Test-time Scaling Techniques in Theoretical Physics - \\
           A Comparison of Methods on the TPBench Dataset}



\icmlsetsymbol{equal}{*}

\begin{icmlauthorlist}
\icmlauthor{Zhiqi Gao}{equal,cs}
\icmlauthor{Tianyi Li}{equal,physics}
\icmlauthor{Yurii Kvasiuk}{physics}
\icmlauthor{Sai Chaitanya Tadepalli}{iub} 
\icmlauthor{Maja Rudolph}{dsi}
\icmlauthor{Daniel J.H. Chung}{physics} 
\icmlauthor{Frederic Sala}{cs}
\icmlauthor{Moritz M\"unchmeyer}{physics}
\end{icmlauthorlist}

\icmlaffiliation{physics}{Department of Physics, University of Wisconsin-Madison}
\icmlaffiliation{cs}{Department of Computer Science, University of Wisconsin-Madison}
\icmlaffiliation{dsi}{Data Science Institute (DSI), University of Wisconsin-Madison}
\icmlaffiliation{iub}{Department of Physics, Indiana University, Bloomington}


\icmlcorrespondingauthor{Frederic Sala}{fredsala@cs.wisc.edu}
\icmlcorrespondingauthor{Moritz M\"unchmeyer}{muenchmeyer@wisc.edu}

\icmlkeywords{Machine Learning, ICML}

\vskip 0.3in
]



\printAffiliationsAndNotice{\icmlEqualContribution} 

\begin{abstract}
Large language models (LLMs) have shown strong capabilities in complex reasoning, and test-time scaling techniques can enhance their performance with comparably low cost. Many of these methods have been developed and evaluated on mathematical reasoning benchmarks such as AIME. This paper investigates whether the lessons learned from these benchmarks generalize to the domain of advanced theoretical physics. We evaluate a range of common test-time scaling methods on the TPBench physics dataset and compare their effectiveness with results on AIME. To better leverage the structure of physics problems, we develop a novel, symbolic weak-verifier framework to improve parallel scaling results. Our empirical results demonstrate that this method significantly outperforms existing test-time scaling approaches on TPBench. We also evaluate our method on AIME, confirming its effectiveness in solving advanced mathematical problems. Our findings highlight the power of step-wise symbolic verification for tackling complex scientific problems.
\end{abstract}

\section{Introduction}
Large language models (LLMs), while increasingly competent at mathematical reasoning, still face challenges with intricate multi-step derivations. To bridge this gap, a variety of test-time scaling techniques have been developed to augment their reasoning capabilities. These techniques, including repeated sampling and self-consistency~\cite{brown2024largelanguagemonkeysscaling}, sampling-plus-verifier pipelines~\cite{zhao2025samplescrutinizescaleeffective,liu20251bllmsurpass405b}, multi-round reflection~\cite{tian2025thinktwiceenhancingllm,muennighoff2025s1simpletesttimescaling}, hierarchical decomposition~\cite{teng2025atomthoughtsmarkovllm,simonds2025ladderselfimprovingllmsrecursive,yang2025reasonfluxhierarchicalllmreasoning}, and RL-based planners~\cite{guan2025rstarmathsmallllmsmaster,pan2025coatchainofassociatedthoughtsframeworkenhancing,hou2025advancinglanguagemodelreasoning}, have shown improvement on standard mathematical benchmarks~\cite{hendrycks2021measuringmathematicalproblemsolving, cobbe2021trainingverifierssolvemath}. This approach presents a distinct computational paradigm from scaling efforts focused on pretraining or extensive finetuning. While test-time compute strategies require more resources than a single inference pass, they can offer a more compute-efficient path to performance gains on complex tasks, challenging traditional scaling laws that often optimize for training costs over inference-time expenditures~\cite{snell2024scalingllmtesttimecompute, bian2025scalinginferenceefficientlanguagemodels}.

Much of the progress in test-time scaling has been developed for mathematical settings. However, theoretical physics reasoning presents distinct challenges compared to undergraduate mathematical level reasoning, such as AIME \cite{AoPS:AIME}, research level mathematical reasoning such as FrontierMath \cite{glazer2024frontiermathbenchmarkevaluatingadvanced}, and proof assistant-based mathematics, where results can be auto-verified \cite{AlphaProofandAlphaGeometryTEams_2024}. For a comparison between the use of math in theoretical physics and in pure math, see \cite{chung2025theoreticalphysicsbenchmarktpbench}. In short, theoretical physics uses direct calculation techniques on mathematical concepts that are not as formally defined as in math.


In this work, we systematically evaluate prominent test-time scaling methods on the TPBench dataset, a collection of advanced physics problems ranging from undergraduate to research level \cite{chung2025theoreticalphysicsbenchmarktpbench}. We find that simple parallel or sequential test time scaling techniques do not perform very well on this dataset. To improve their performance, we develop a novel weak-verifier framework that uses a SymPy-augmented agent to verify the mathematical steps within each candidate solution, providing a robust mechanism for identifying correct reasoning paths. 
Our contributions are:
\begin{itemize}
    \item A comprehensive empirical study of existing test-time scaling techniques, including sampling-based search and multi-round reasoning, on the TPBench theoretical physics dataset. Our analysis investigates how the effectiveness of these methods compares between physics and mathematical reasoning domains.
    \item A novel step-wise weak-verifier pipeline that integrates symbolic computation (SymPy) to grade and select solutions. Our results show this pipeline surpasses other test-time scaling methods by a significant margin on TPBench, achieving up to a 22\% improvement and approaching the theoretical Best-of-N performance. We also test our verifier on the AIME benchmark \cite{AoPS:AIME}, confirming its general effectiveness.
\end{itemize}

Our work highlights the importance of tailoring inference-time strategies to specific scientific domains and demonstrates the power of symbolic verification in various domains.

\section{Related Work}

\paragraph{Test-Time Scaling for Reasoning.}
Large language models have recently shown that increasing inference-time computation---often called test-time scaling---can significantly boost reasoning performance \cite{snell2024scalingllmtesttimecompute, bian2025scalinginferenceefficientlanguagemodels}.  OpenAI's proprietary o1 model exemplified this by using extended ``slow thinking'' procedures to achieve stronger results than standard one-shot prompting. This has spurred a wave of research into inference-time strategies that improve mathematical problem solving without changing model weights.

\paragraph{Sampling-Based Search and Verification.}
A common way to leverage more compute at test time is to sample multiple responses in parallel and choose a final answer by some criterion. This approach uses the randomness of LLMs. Brown et al.\ show that repeated sampling at large scales steadily increases the probability of including a correct solution at least once, boosting solve rates on code and formal proof tasks dramatically \cite{brown2024largelanguagemonkeysscaling}. Zhao et al.\ extend this idea with a ``sample, scrutinize and scale'' method: generate hundreds of candidate solutions and use the LLM itself as a verifier to score and select the best one \cite{zhao2025samplescrutinizescaleeffective}. Liu et al.\ rethink the allocation of inference budget between a solution-proposing policy model and a process reward model, demonstrating that small models with optimal compute splits can rival or beat much larger ones on challenging math benchmarks \cite{liu20251bllmsurpass405b}.

\paragraph{Iterative and Multi-Round Reasoning.}
As a comparison to parallel sampling, sequential approaches such as iterative refinement let the model revisit its answer through multiple rounds. Tian et al.\ propose \emph{Think Twice}, which feeds the model's initial answer back into itself for a re-answer step, yielding consistent gains on AIME and other math competition tasks \cite{tian2025thinktwiceenhancingllm}. Muennighoff et al.\ introduce \emph{s1}, using ``budget forcing'' to extend chain-of-thought generation (``Wait'' tokens) and fine-tuning on high-quality reasoning traces, force the answer to have a longer CoT and a better performance \cite{muennighoff2025s1simpletesttimescaling}. Wang et al.\ observe that models often underthink---prematurely abandoning reasoning paths---and propose a Thought Importance Penalty decoding tweak to encourage deeper exploration of each line of thought \cite{wang2025thoughtsplaceunderthinkingo1like}. Yang et al.\ develop \emph{Step Back to Leap Forward}, endowing the LLM with self-backtracking capabilities to undo and retry flawed steps to enhance mathematical reasoning capabilities. \cite{yang2025stepleapforwardselfbacktracking}.

\paragraph{Hierarchical and Decompositional Strategies.}
Breaking complex problems into smaller subtasks is another effective paradigm. Teng et al.\ introduce \emph{Atom of Thoughts} (AoT), treating reasoning as a Markov chain of atomic sub-questions that are solved independently and contracted back into the main problem, preventing distraction by irrelevant context \cite{teng2025atomthoughtsmarkovllm}. Simonds and Yoshiyama's \emph{LADDER} framework has models that recursively generate simpler variants of a hard problem, solve them, and use these solutions to tackle the original, achieving dramatic gains on integration calculus questions \cite{simonds2025ladderselfimprovingllmsrecursive}. Yang et al.'s \emph{ReasonFlux} equips a model with a library of high-level thought templates and trains a hierarchical planner via reinforcement learning to assemble them, pushing state-of-the-art results on MATH and AIME benchmarks \cite{yang2025reasonfluxhierarchicalllmreasoning}.

\paragraph{Search-Based Planning and RL-Enhanced Inference.}
Several works combine LLMs with classical search or reinforcement learning at inference. Guan et al.'s \emph{rStar-Math} pairs a policy model with a process reward model under a Monte Carlo Tree Search framework, enabling a 7B model to match or exceed much larger counterparts on MATH and AIME without distillation \cite{guan2025rstarmathsmallllmsmaster}. Pan et al.'s \emph{CoAT} (Chain-of-Associated-Thoughts) uses MCTS with an associative memory mechanism for dynamic knowledge integration during the search, improving solution coherence and diversity \cite{pan2025coatchainofassociatedthoughtsframeworkenhancing}. Hou et al.\ train a 32B model via reinforcement learning from synthetic and human-feedback reasoning traces to encourage ``slow thinking'' behaviors, yielding smooth inference-scaling improvements without external verifiers \cite{hou2025advancinglanguagemodelreasoning}.

\paragraph{TPBench.}
Theoretical Physics Benchmark (TPBench) \cite{chung2025theoreticalphysicsbenchmarktpbench} is a new benchmark dataset designed to evaluate AI's ability to solve high-level theoretical physics problems, specifically in high-energy theory and cosmology. This benchmark addresses the gap in existing datasets, which primarily focus on high-school-level physics, by including 57 novel problems ranging from undergraduate to research level. The paper also discusses the unique reasoning required for theoretical physics, contrasting it with mathematical reasoning, and highlights the potential of AI as a valuable tool for theoretical physicists by assisting with complex calculations and potentially contributing to the generation of new research ideas.

\section{Methodology} 

In this section, we describe the generation of samples, our sequential scaling approach, and several variations of the parallel scaling approach, including a SymPy augmented verifier.

\subsection{Generation and Evaluation of Solutions}

We mainly use Gemini models for evaluation in this paper. Gemini-2.5-Pro achieves the best performance across all models, producing the best result we can obtain via test-time scaling methods for current state-of-the-art models. Gemini-2.0-Flash is a cheap and fast model with relatively strong performance. To show that our results are consistent across model families, we also use o4-mini-high as an alternative model for testing.

We generate solutions using the prompt in Appendix \ref{app:prompt_default}. To evaluate solutions, we implemented an auto-verification pipeline to automate solution assessment as described in the TPBench paper \cite{chung2025theoreticalphysicsbenchmarktpbench}. This system requires problem solutions to be submitted as Python callables with predefined signatures in each round of generation. 




\subsection{Sequential: Multi-Round Reasoning}
This methodology uses a multi-round reasoning strategy to iteratively refine the problem-solving process. This is repeated for a fixed number of iterations, allowing the model to build upon or reconsider its previous lines of thought. To limit context length in multi-round reasoning, we use a summarization technique at each iteration of the process.

The pseudo-code is shown below:

\begin{algorithm}[H] 
\caption{\text{Multi-Round Reasoning and Solution Generation}}
\label{alg:simplified_multi_round_reasoning}
\begin{algorithmic}[1] 
    \REQUIRE \text{General contextual instructions ($GCI$)} \text{Problem statement ($PS$)} \text{Specific Python code requirements ($SCR$)} \text{Number of iterations ($N_{iter}$)}
    \FOR{$i \leftarrow 1$ \textbf{to} $N_{iter}$}
        \IF{$\text{PreviousThinking}$ is Null}
            \STATE $\text{Reasoning} \leftarrow \text{LLM}(\text{GCI}, \text{PS})$
        \ELSE
            \STATE $\text{Reasoning} \leftarrow \text{LLM}(\text{GCI}, \text{PS}, \text{PreviousThinking})$
        \ENDIF
        \STATE $\text{PreviousThinking} += \text{Reasoning}$

        \STATE $\text{Code Answer(for evaluation purpose)} \leftarrow \text{LLM}(\text{Reasoning steps above, SCR})$

    \ENDFOR
\end{algorithmic}
\end{algorithm}

In each iteration, the core task is to generate step-by-step reasoning towards a solution. For the initial iteration, the language model is provided with general contextual instructions and the specific problem statement, guiding it to begin its reasoning using the initial prompt in Appendix \ref{app:multi-round prompt}. For all subsequent iterations, the process encourages a re-evaluation or continuation of prior efforts. The prompt, which is the subsequent iteration prompt \ref{app:multi-round prompt}, incorporates the original problem statement and general instructions, along with the summarization by the process above.

Following the generation of detailed reasoning, a separate step is taken to synthesize the results into a final symbolic mathematical expression and then translate that expression into executable Python code (using the Specific Python code requirements), which is needed for the TPBench auto-verification process (step 8 in algorithm \ref{alg:simplified_multi_round_reasoning}). This synthesis is guided by a structured prompt that incorporates the accumulated reasoning and any specific constraints or formatting requirements for the Python code by the summarization prompt in \ref{app:multi-round prompt}. Finally, we evaluate correctness using our pipeline and the generated Python code.

\subsection{Parallel: Majority Vote}
\label{subsec:generation}
For all of our parallel methods, we generate $N=50$ candidate solutions denoted $\mathcal{S}=\{s_1,\dots,s_{50}\}$ to decrease randomness. 

In the Majority Vote strategy, each of the $N=50$ candidate solutions is first evaluated numerically as discussed in the auto-verification pipeline. The final answer is determined by identifying the numerical solution tuples that appear most frequently among all candidates. Finally, we check the correctness of the voted answer and mark the problem as solved if the voted answer is correct.

\subsection{Parallel: Best of N}
For the Best of N strategy, all $N=50$ candidate solutions 
$i \in S$ are individually assessed for correctness using the auto-verification pipeline. If at least one candidate solution $i$ is verified as correct by the pipeline, we mark this problem as solved. This is used as the theoretical upper bound for all sampling-based strategies.

\subsection{Parallel: Sampling-Based Search with Weak Verifiers}
\label{subsec:parallel_sampling_verifiers}
The sampling-based search methods we discuss adapt the general verification strategies proposed by Zhao\,\textit{et al.}\,\cite{zhao2025samplescrutinizescaleeffective}. This common framework involves generating multiple candidate solutions and then employing a verifier to score and select the best one. We implement this framework using two distinct approaches for Stage 2: alongside a \textbf{simple weak verifier}, we introduce a \textbf{symbolic verifier augmented strategy}, specifically designed for physics problems, which leverages a SymPy-based agent. Thus, the core difference between our methods lies in the mechanism of the verifier used in Stage 2. Here are the details of this framework. The overall process is described in Algorithm~\ref{alg:sampling_search_revised}.

\paragraph{Stage~1: Identification of Functionally Distinct Solutions.}
From the initial $N=50$ candidate solutions, $\mathcal{S}=\{s_1,\dots,s_{50}\}$ (generated as per Section~\ref{subsec:generation}), we first identify a subset of functionally distinct attempts. This process leverages an auto-verification framework where the final mathematical expression derived from each candidate solution $s_i$ is translated into a programmatic function, denoted $f_i$. Each such function $f_i$ is then systematically evaluated against a predefined suite of $M=5$ distinct test case inputs $(x_1, \dots, x_M)$. This evaluation yields an output vector $\mathbf{o}_i = (f_i(x_1), \dots, f_i(x_M))$ for each solution $s_i$. Two solutions, $s_i$ and $s_j$, are deemed functionally distinct if their respective output vectors $\mathbf{o}_i$ and $\mathbf{o}_j$ are not identical (i.e., $\mathbf{o}_i \neq \mathbf{o}_j$). Solutions that produce identical output vectors across all $M$ test cases are considered functionally equivalent. By selecting one representative from each group of functionally equivalent solutions, we derive a pruned set $\mathcal{S}' \subseteq \mathcal{S}$ containing only functionally distinct candidates for subsequent evaluation stages. This set $\mathcal{S}'$ is used as the input $S$ in Algorithm~\ref{alg:sampling_search_revised}.

\paragraph{Stage~2: Solution Verification and Scoring.}
For every candidate solution $s_i \in \mathcal{S}'$, a verification score $\bar{V}_i$ is computed. This stage differentiates the two approaches:

\noindent \textit{Simple Weak Verifier Strategy.}
In this approach, the language model ($LM$) itself acts as a simple verifier. For each candidate solution $s_i$, the $LM$ is queried $k_{\mathrm{verif}}$ times (here 10) with a prompt asking whether solution $s_i$ is correct for the given problem $Q$. Each query $j$ yields a binary score $V_{ij} \in \{0, 1\}$ (1 for correct, 0 for incorrect). The average score $\bar{V}_i = \frac{1}{k_{\mathrm{verif}}}\sum_{j=1}^{k_{\mathrm{verif}}} V_{ij}$ is then used.

\noindent \textit{Symbolic Verifier Augmented Strategy.}
This strategy employs a more sophisticated, physics-augmented LLM-based grading agent for verification. For every candidate solution $s_i \in \mathcal{S}'$, this specialized agent conducts a meticulous, step-by-step evaluation. A key feature is its capability for symbolic computation: for each identified mathematical calculation step within the solution's derivation, it automatically generates and executes a Python script leveraging the SymPy library to independently verify the computation. The agent then compares the result from SymPy with the corresponding step in the solution $s_i$. Based on this comprehensive verification, which encompasses both the logical flow and the correctness of individual calculations confirmed by SymPy, the agent assigns a binary score $V_{ij} \in \{0, 1\}$ to the solution $s_i$ for each of the $k_{\mathrm{verif}}$ verification repetitions. The average verification score is $\bar{V}_i = \frac{1}{k_{\mathrm{verif}}}\sum_{j=1}^{k_{\mathrm{verif}}} V_{ij}$.

High-quality solutions are then retained in a set $\mathcal{S}_{\mathrm{best}} = \Bigl\{ s_i \in \mathcal{S}' \;\Big|\; \bar{V}_i \ge \max_{s_j \in \mathcal{S}'} \bar{V}_j - \delta \Bigr\}$, where $\delta=0.05$ is a tolerance parameter, particularly for the symbolic verifier strategy (for the simple verifier, $\delta$ can be set to 0 to select only top-scoring candidates or adjusted as needed).

\paragraph{Stage~3: Pairwise Tie‑Breaking.}
If Stage 2 results in $|\mathcal{S}_{\mathrm{best}}|>1$ (i.e., multiple solutions are deemed high-quality or are tied), pairwise comparisons are performed as detailed in Algorithm~\ref{alg:sampling_search_revised}, lines 14-20. For each unordered pair $(s_a,s_b)\subset\mathcal{S}_{\mathrm{best}}$, the LLM is prompted $k_{\mathrm{tie}}$ times:
\begin{quote}\small
\texttt{Given problem Q, which solution is more correct?\\
A) $s_a$ (plausibility $H_p(s_a)$)\\
B) $s_b$ (plausibility $H_p(s_b)$)}
\end{quote}
The solution winning the majority of these $k_{\mathrm{tie}}$ comparisons for a given pair is favored in that matchup. The final answer $s^\star$ is the solution that wins the most matchups overall.

\begin{algorithm}[h] 
\caption{Sampling-Based Search with Weak Verifiers}
\label{alg:sampling_search_revised}
\begin{algorithmic}[1] 
    \REQUIRE language model $LM$, SymPy agent, problem $Q$, verification count $k_{\mathrm{verif}}$, verification strategy $VS$
    \STATE Generate 50 samples by the generation method in Section~\ref{subsec:generation}.
    \STATE Select distinct solution set $S$.
    \FOR{Each candidate response $s_i \in S$}
        \IF{$VS = \text{Simple}$}
            \STATE For $s_i$: Score $\bar{V}_i \leftarrow $ fraction of ``correct" (1s) from $LM$ over $k_{\mathrm{verif}}$ queries for problem $Q$.
        \ELSIF{$VS = \text{Symbolic}$}
            \STATE For $s_i$: Score $\bar{V}_i \leftarrow $ fraction of ``correct" (1s) from SymPy agent over $k_{\mathrm{verif}}$ checks for problem $Q$.
        \ENDIF
    \ENDFOR
    \STATE Let $V_{\max} = \max_{s_k \in S} \{\bar{V}_k\}$.
    \STATE Gather the highest-scored responses $S_{\text{Best}} = \{s_i \in S \mid \bar{V}_i = V_{\max}\}$.
    \IF{$|S_{\text{Best}}| = 1$}
        \STATE Let $s_{i^*}$ be the single response in $S_{\text{Best}}$. Return $s_{i^*}$.
    \ELSE
        \FOR{each pair of $(s_i, s_j) \in \binom{S_{\text{Best}}}{2}$}
            \STATE $C_{i,j} \leftarrow LM(\text{Evaluate } s_i \text{ vs. } s_j \text{ for } Q)$

        \ENDFOR
        \STATE Return response $s_{i^*}$ where $i^*$ corresponds to the $s_i \in S_{\text{Best}}$ that is selected most often in $\{C_{i,j}\}$.
    \ENDIF
\end{algorithmic}
\end{algorithm}

\paragraph{Implementation using Agent Framework (for Symbolic Verifier).}
The SymPy-augmented LLM verification process (used in Stage 2 for the Symbolic Verifier Augmented Strategy) is implemented using the \texttt{openai-agents} API. While our implementation relies on this particular framework, a number of other agent-based libraries could be used equivalently to achieve the same functionality. The system employs an LLM-based grading agent governed by a detailed set of instructions (provided in Appendix~\ref{app:physics_grader_prompt}) and equipped with a list of tools. The key tool for this verification is a custom Python function, \texttt{run\_sympy\_script}. This tool is designed to execute SymPy scripts: the agent identifies mathematical calculation steps within a candidate solution's derivation, formulates a corresponding SymPy script for each, and then invokes the \texttt{run\_sympy\_script} tool. The tool executes the provided script in a sandboxed environment, captures its standard output (STDOUT) and standard error (STDERR), and returns these to the agent. The agent subsequently compares this SymPy output with the assertion made in the solution's step to verify its correctness. This iterative, tool-mediated verification enables a meticulous, step-by-step assessment of the mathematical reasoning, with all findings reported in a structured JSON format alongside an overall binary score for that verification run ($V_{ij}$), as detailed in Appendix~\ref{app:physics_grader_prompt}.

\section{Experiments}
\subsection{Experiment Setup}
\textbf{TPBench}. We use problems from level 3 to level 5, which is graduate and research-level problems in the datasets, to test our hypothesis. There are 11, 14, and 11 problems, respectively, in levels 3 to 5.

\textbf{AIME}. As AIME problem numbers are roughly sorted by difficulty, we select problems numbered 11-15 from AIME 2024 and 2025 to create a 20-problem set.

\subsection{Answer Distribution Analysis}
\begin{figure*}[ht]
\vskip 0.2in
\begin{center}
\centerline{\includegraphics[width=\textwidth]{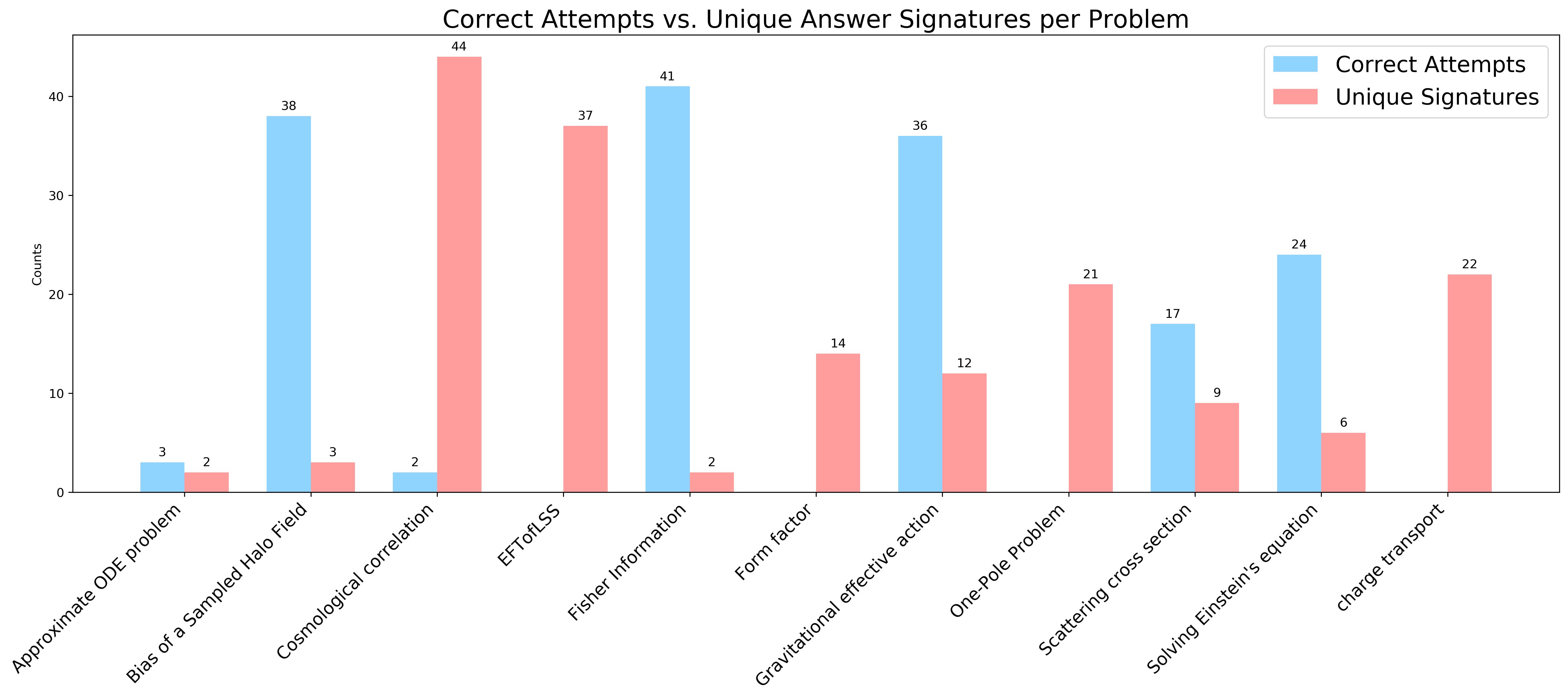}}
\caption{Analysis of 50 attempts per Level 5 TPBench problem by Gemini-2.5 Pro. The red bars show the total number of unique attempts (i.e., inequivalent answers) generated for each problem, while the blue bars indicate the number of correct attempts among those 50 samples. This visualization highlights the raw solution space from which our weak verifier selects.}
\label{fig:attempts_analysis}
\end{center}
\vskip -0.2in
\end{figure*}

To understand the potential of sampling-based approaches, we first analyzed the distribution of solutions generated by repeatedly sampling the model. For each Level 5 problem in our test set, we generated $N=50$ attempts. Figure \ref{fig:attempts_analysis} illustrates the outcome of this initial sampling stage.

As depicted in Figure \ref{fig:attempts_analysis}, there is considerable variation across problems in both the number of unique solutions and the number of correct solutions found within the 50 attempts. For some problems, a significant fraction of the attempts are correct, providing a rich pool for a verifier. For others, correct solutions are sparse. Notably, the number of unique attempts (blue bars) is often less than 50, indicating that the model frequently generates identical or semantically equivalent incorrect (or correct) reasoning paths. This redundancy underscores the importance of filtering for distinct solutions, as proposed in our methodology.

The key insight from this initial analysis is that even with a fixed number of 50 samples, a correct solution is often present within the generated set. The challenge, which our weak verifier aims to address, is to effectively identify these correct instances from the typically larger set of unique (but not necessarily correct) attempts. The orange bars represent the best-case scenario for a perfect verifier operating on these 50 samples. Our approach aims to bridge the gap between raw sampling and this "oracle" best-of-$N$ performance by leveraging domain-specific checks on the unique solutions.

\subsection{Experimental Results}

We benchmark the various methods on the TPBench data set for difficulties 3 to 5 (difficulties 1 and 2 are too easy for the Gemini family). The results are summarized in Tables \ref{table:tpbench_gemini25} (Gemini 2.5 pro), \ref{table:tpbench_gemini20} (Gemini 2.0 flash) and \ref{table:tpbench_o4mini}. Each table contains the average accuracy in a single attempt as a baseline. Note that both weak verifiers use the same model as the one employed to generate the solutions.

\begin{table}[h!]
    \centering
    \caption{Comparison of test-time scaling approaches on TPBench With Gemini 2.5 Pro.}
    \label{tab:method_comparison_booktabs_only}
    \begin{tabular}{lrr}
        \toprule
        Method & Level 4 & Level 5 \\
        \midrule
        Single Attempt & 63.3\% & 29.3\% \\
        \midrule
        \multicolumn{3}{l}{\textit{Sequential Methods}} \\
        1+Round Reasoning & 65.0\% & 26.4\% \\
        2+Round Reasoning & 65.0\% & 26.4\% \\
        4+Round Reasoning & 68.6\% & 28.2\% \\
        \midrule
        \multicolumn{3}{l}{\textit{Parallel Methods}} \\
                Simple Weak Verifier & 71.4\% & 27.3\% \\

        Majority Vote & 78.6\% & 36.4\% \\
        SymPy Verifier & 71.4\% & 54.5\% \\
        Best of N & 85.7\% & 63.6\% \\
        \bottomrule
    \end{tabular}
    \par
    \small \textit{Note:} Single attempt's accuracy is averaged over 50 attempts; multi-round reasoning over 10 attempts; parallel methods use 50 candidates.
    \label{table:tpbench_gemini25}
\end{table}

\begin{table}[h!]
    \centering
    \caption{Comparison of test-time scaling approaches on TPBench With Gemini 2.0 Flash}
    \label{tab:method_comparison_vertical_improved}
    \begin{tabular}{lccc}
        \toprule
        Method & Level 3 & Level 4 & Level 5 \\
        \midrule
        \multicolumn{4}{l}{\textit{Baseline}} \\
        Single Attempt & 52.5\% & 13.1\% & 1.5\% \\
        \midrule
        \multicolumn{4}{l}{\textit{Sequential Methods}} \\
        1+Round & 61.8\% & 16.4\% & 2.7\% \\
        2+Round & 60.0\% & 22.9\% & 0.9\% \\
        4+Round & 60.0\% & 21.4\% & 1.8\% \\
        \midrule
        \multicolumn{4}{l}{\textit{Parallel Methods}} \\
        Simple Weak Verifier & 18.2\% & 7.1\% & 0\% \\
        Majority Vote & 72.7\% & 35.7\% & 9.1\% \\
        SymPy Verifier & 81.8\% & 57.1\% & 9.1\% \\
        Best of N & 90.9\% & 85.7\% & 18.2\% \\
        \bottomrule
    \end{tabular}
    \label{table:tpbench_gemini20}
\end{table}

\begin{table}[h!]
    \centering
    \caption{Comparison of test-time scaling approaches on TPBench with o4-mini-high}
    \label{tab:tpbench_o4_mini}
    \begin{tabular}{lrr}
        \toprule
        Method & Level 4 & Level 5 \\
        \midrule
        \multicolumn{3}{l}{\textit{Baseline}} \\
        Single Attempt & 53.0\% & 21.1\% \\    
        \midrule
        \multicolumn{3}{l}{\textit{Sequential Methods}} \\
        1+Round Reasoning & 50.0\% & 19.1\% \\
        2+Round Reasoning & 52.9\% & 18.2\% \\
        4+Round Reasoning & 52.9\% & 20.0\% \\
        \midrule
        \multicolumn{3}{l}{\textit{Parallel Methods}} \\
                Simple Weak Verifier & 64.3\% & 18.2\% \\

        Majority Vote & 71.4\% & 27.3\% \\
        SymPy Verifier & 71.4\% & 36.4\% \\
        Best of N & 78.6\% & 54.5\% \\
        \bottomrule
    \end{tabular}
    \label{table:tpbench_o4mini}
\end{table}

\begin{table}[h!]
    \centering
    \caption{Comparison of test-time scaling approaches on AIME-Subset}
    \label{tab:aime_subset_comparison}
    \begin{tabular}{lrr}
        \toprule
        Model & Gemini-2.0 & Gemini-2.5 \\
        \midrule
                \multicolumn{3}{l}{\textit{Baseline}} \\

        Single Attempt & 11.4\% & 71.6\% \\
        \midrule
        \multicolumn{3}{l}{\textit{Sequential Methods}} \\
        1+Round Reasoning &  8.5\% & 73.0\% \\
        2+Round Reasoning &  8.5\% & 73.0\% \\
        4+Round Reasoning &  8.5\% & 73.0\% \\
        \midrule
        \multicolumn{3}{l}{\textit{Parallel Methods}} \\
                Simple Weak Verifier & 20\% & 60\% \\

        Majority Vote & 15\% & 80\% \\
        SymPy Verifier & 25\% & 80\% \\
        Best of N & 35\% & 100\% \\
        \bottomrule
    \end{tabular}
    \label{table:aime_gemini25_20}
\end{table}

\textbf{Simple parallel scaling.} The natural baseline approach to parallel scaling is majority vote. We find that the majority vote is generally a good way for selecting the correct answer when we have a relative large answer pool. The second method we explore to select the answer is LLM self-grading, called the Simple Weak Verifier Strategy above. We find that this method does not improve results even over the average single accuracy. The reason for the failure mainly originates from the low precision in the grading procedure. We also found that the tie-breaking stage has near-random success. This is in line with findings that LLMs are not good at spotting their own mistakes \cite{huang2024largelanguagemodelsselfcorrect, stechly2024selfverificationlimitationslargelanguage}. 

\textbf{Parallel scaling with SymPy pipeline.} We find strong improvments using our agent-based step-by-step symbolic pipeline. For example, with Gemini 2.5 pro, we go from $29.3 \%$ average accuracy to $54.4 \%$, close to the best-of-N upper bound of $63.6 \%$. We note that this pipeline performs a step-by-step verification. This means that gains in performance come both from re-visiting individual steps with an LLM and from writing Sympy code for individual steps where possible. More details about this verifier are given in Sec. \ref{sec:symbverif}, and two examples are shown in App. \ref{app:halo_bias_verification} and  \ref{app:one_pole_verification}.

\textbf{Sequential scaling.} We find that sequential scaling only leads to a minor improvement of model performance, and in some cases even to a decrease in performance. Further, there is no clear improvement with additional rounds of reasoning. Our method, loosely inspired by the s1 approach \cite{muennighoff2025s1simpletesttimescaling}, does not work well for this data set. We plan to examine sequential scaling in more detail in the future. 

\textbf{Comparison with AIME.} We compare the same methods on our subset of AIME problems in Table \ref{table:aime_gemini25_20}. We find similar improvements to those in the TPBench case. However, in this case, the Sympy verifier is identical in performance to the Majority Vote. This could indicate that symbolic verification is more important in theoretical physics, which includes more complicated calculations (compared to AIME, not necessarily math in general).

\subsection{Scaling of Parallel Methods}

\begin{figure}[ht]
    \centering
    \includegraphics[width=\linewidth]{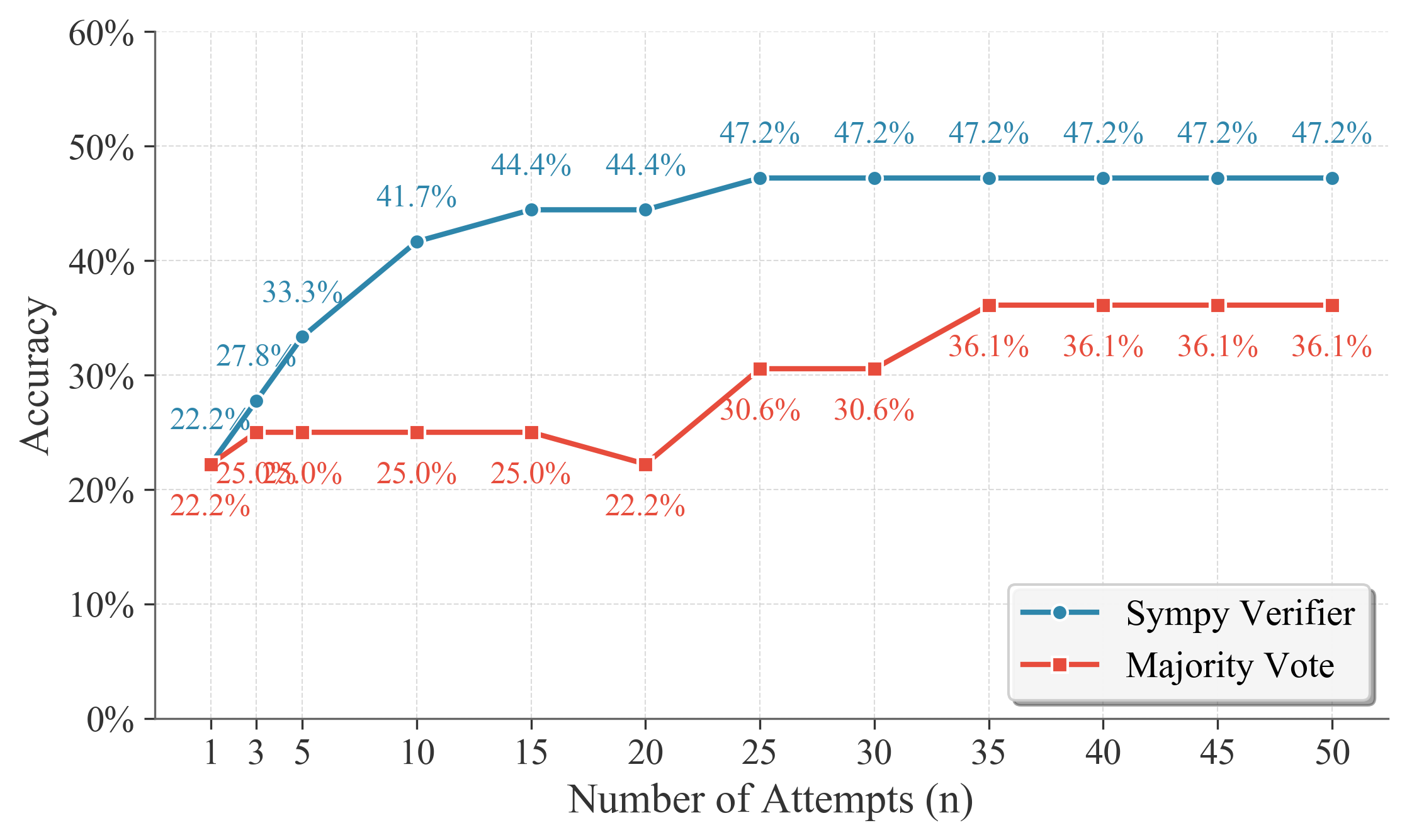}
    \caption{Scaling of accuracy with the number of parallel attempts ($N$) for Majority Vote and our proposed SymPy-augmented weak-verifier framework. Data is from TPBench problems of difficulty levels 3, 4, and 5, using Gemini 2.0 Flash. The plot illustrates the improvement in solution accuracy as more candidate solutions are generated and evaluated by each method.}
    \label{fig:scaling_parallel_attempts}
\end{figure}

To investigate the scalability of parallel test-time scaling approaches, we analyzed the performance of Majority Vote and our SymPy-augmented weak-verifier framework as a function of the number of generated candidate solutions ($N$). Figure~\ref{fig:scaling_parallel_attempts} illustrates the accuracy achieved by these two methods, across different problem difficulty levels (Level 3, 4, and 5) from TPBench, using Gemini 2.0 Flash as the base model.

The results show a consistent trend for both evaluated methods: accuracy improves with an increasing number of attempts for all difficulty levels. The SymPy-augmented verifier consistently outperforms Majority Vote across all tested $N$ values, and saturates at a higher value. It appears that with 50 samples, we are approaching a saturation point, indicating a limit on the maximum performance one can reach with these models and approaches. This aligns with the finding from Wu et al. that shows accuracy from inference flops, the majority voting and weighted majority voting methods follows a curve that saturates. \cite{wu2025inferencescalinglawsempirical}

\subsection{Scaling of Sequential Method}

\begin{figure}[H]
    \centering
        \includegraphics[width=\linewidth]{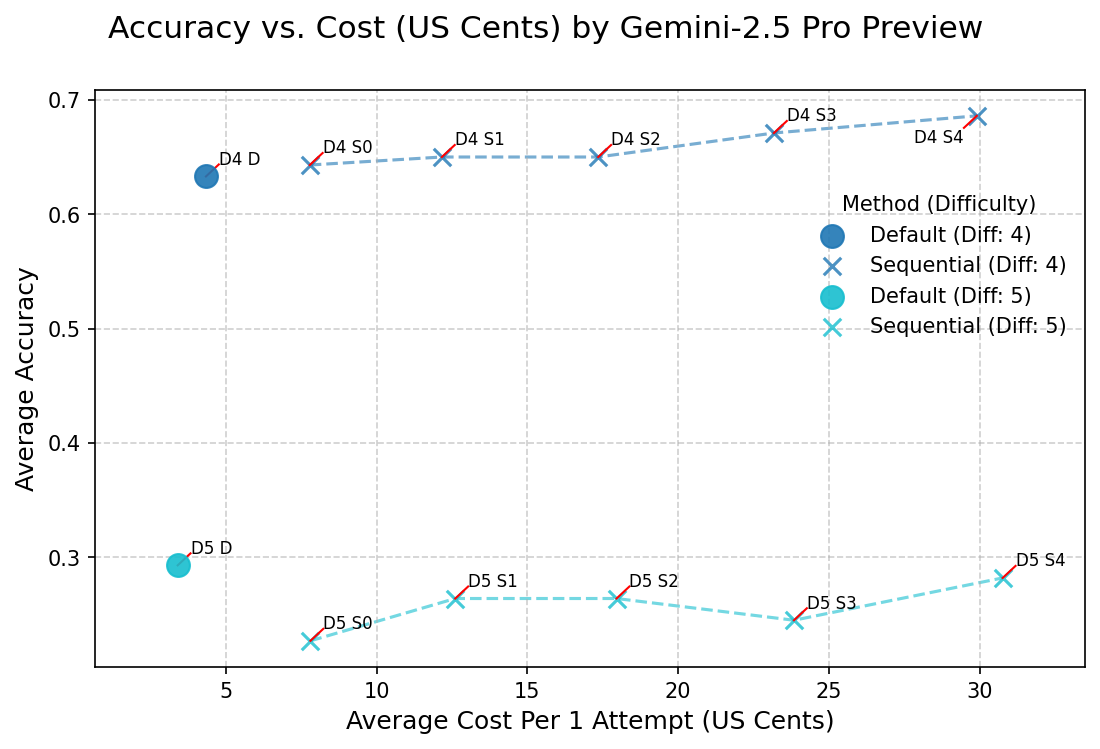}
\end{figure}
\begin{figure}[H]
    \centering
    \includegraphics[width=\linewidth]{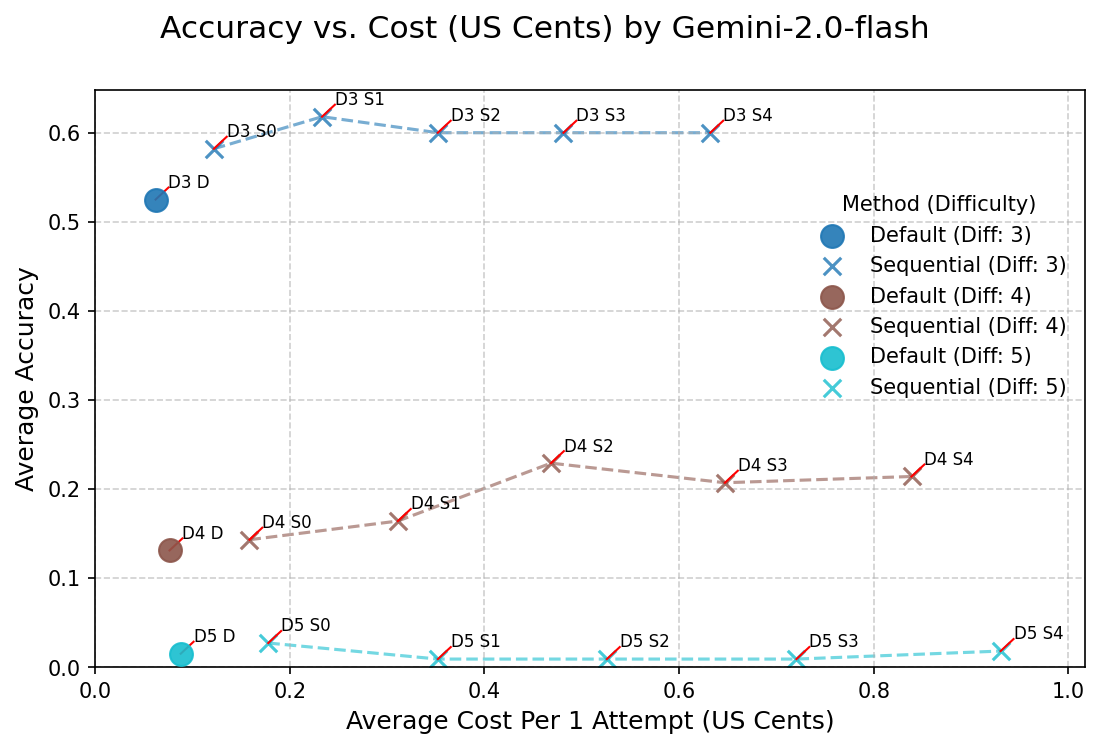}
    \caption{Cost and accuracy comparison for single attempt and sequential scaling method.}
    \label{fig:costvsaccuracy}
\end{figure}

We also examined the computational cost of sequential scaling compared to the achieved accuracy. In figure \ref{fig:costvsaccuracy}, the dot $DnSi$ indicate the average cost for a problem with difficulty level $n$ and in the round $i$ in sequential scaling, where round $i$ implies the number of extra thinking rounds the model takes. The detailed token usage is illustrated in Appendix \ref{app:token_sequential}. Since we have an extra summarization step to generate the evaluation code, the cost of $DnS0$ is higher than the cost for a single attempt than the default reasoning method. In terms of performance, we see that we get a minor improvement on questions with difficulty levels 3 and 4, but no or negative improvement on level 5 for both models. This aligns with the finding in the S1 paper \cite{muennighoff2025s1simpletesttimescaling}, where their s1 and s1.1 models almost have no improvement for appending more ``wait" tokens on the hard AIME task, but have minor improvement on easier MATH500 tasks. This shows the limit of the sequential scaling method on hard problems.

\subsection{Performance of the Symbolic Verifier}
\label{sec:symbverif}


While our SymPy-augmented weak verifier outperforms other methods, it is crucial to analyze its performance critically. The verifier's score represents an automated judgment, not an infallible ground truth. This section dissects the verifier's operational statistics, quantifies its accuracy against a manually graded baseline, and outlines its current capabilities and limitations, framing it as a promising but foundational step toward more robust automated scientific reasoning.

\paragraph{Verifier Usage Statistics.}
First, we examine how the symbolic verifier operates during the evaluation of candidate solutions. As shown in Table~\ref{tab:verifier_usage_stats}, the agent's reasoning process is heavily reliant on symbolic computation. On average, each candidate solution is broken down into approximately 8 distinct logical or mathematical steps. Of these, a remarkable 83\% involve the generation and execution of a SymPy script, confirming that the agent actively uses its symbolic tool to validate calculations rather than relying solely on its internal knowledge. Furthermore, the tie-breaker stage was activated in 31\% of the problems, indicating that it is common for multiple, functionally distinct solutions to receive a perfect score from the verifier, making the pairwise comparison a necessary final step for disambiguation.

\begin{table}[h!]
\centering
\caption{Operational statistics of the SymPy-augmented verifier on Level 4 and 5 problems. The data shows a deep reliance on symbolic checks and frequent use of the tie-breaking mechanism.}
\label{tab:verifier_usage_stats}
\begin{tabular}{lc}
\toprule
Metric & Value \\
\midrule
Avg. verification steps per solution & 8.2 \\
Verification steps with SymPy script & 83\% \\
Agent-perceived correctness of steps & 88\% \\
Problems requiring tie-breaker & 31\% \\
\bottomrule
\end{tabular}
\end{table}

\paragraph{Capabilities and Limitations.}
The verifier's strength lies in its ability to catch concrete algebraic and calculus errors. For instance, in the cosmological halo bias problem detailed in Appendix~\ref{app:halo_bias_verification}, the verifier successfully identified errors in the calculation of truncated Gaussian moments. By symbolically re-computing the integrals, it flagged that the candidate's first moment calculation was incorrect by a factor of the variance $\sigma$, a subtle but critical mistake that invalidated the final result. 
However, the verifier's capabilities are bound by the scope of its SymPy toolset. As summarized in Table~\ref{tab:verifier_capabilities}, it struggles with mathematics that is highly abstract or lacks a straightforward algorithmic representation.

\begin{table}[h!]
\centering
\caption{A summary of the symbolic verifier's current capabilities for different types of mathematical operations.}
\label{tab:verifier_capabilities}
\begin{tabular}{p{5cm}c}
\toprule
Mathematical Operation & Verifiable? \\
\midrule
Polynomial \& Rational Function Algebra & Yes \\
Standard Differentiation \& Integration & Yes \\
Residue Calculation at Poles & Yes \\
Approximations \& Limit-taking & Partially \\
General Tensor Manipulations (GR) & No \\
Advanced Path Integrals (QFT) & No \\
\bottomrule
\end{tabular}
\end{table}

The agent cannot, for example, robustly verify steps involving tensor calculus in arbitrary spacetimes or abstract path integrals, as these concepts lack direct, general-purpose implementations in SymPy. This limitation explains some of the recall failures, where correct but advanced reasoning steps were flagged as unverifiable and thus penalized.

Our analysis shows that symbolic verification is a powerful and necessary tool for advancing AI in theoretical physics. It provides a substantial boost in identifying correct solutions by grounding the LLM's reasoning in formal computation. Nonetheless, its imperfections and the clear boundaries of its capabilities underscore that this is merely a first step. Future work must focus on expanding the verifier's toolkit and improving its ability to parse and validate more abstract forms of physical and mathematical reasoning.

\section{Discussion \& Conclusion}
In this paper, we address the challenge of enhancing the reasoning capabilities of large language models (LLMs) for advanced theoretical physics problems, particularly where existing test-time scaling techniques have shown limited success.  We conducted a comprehensive empirical study of sampling-based parallel scaling and multi-round reasoning on the TPBench, a benchmark for advanced physics problems up to the research level, as well as on AIME. In general, we find that parallel methods work well, in particular when augmented with a step-wise symbolic verification procedure.  In future work, we aim to build stronger domain-specific verification tools, for example, for tensor operations in general relativity. It would also be important to use symbolic tools at the time of solution generation rather than only for verification. For sequential scaling, we do not find improvements on hard problems. Improving sequential scaling on theoretical physics problems is a further interesting direction for future research. Finally, parallel and sequential techniques could be combined to compute optimal ways.



\section*{Acknowledgements}

M.M. and D.J.H.C. acknowledge the support by the U.S. Department of Energy, Office of Science, Office of High Energy Physics under Award Number DE-SC0017647. M.M. also acknowledges the support by the National Science Foundation (NSF) under Grant Number 2307109 and
the Wisconsin Alumni Research Foundation (WARF). F.S. is grateful for the support of the NSF under CCF2106707 and
the Wisconsin Alumni Research Foundation (WARF).

\bibliography{paper_test_time_scaling_arxiv/references}
\bibliographystyle{icml2025}

\nocite{langley00}

\newpage
\appendix
\onecolumn

\section{All prompts used in this paper}
\subsection{Prompt for default generating method}
\label{app:prompt_default}
\begin{lstlisting}[
    basicstyle=\ttfamily\scriptsize,
    frame=single,
    framerule=0.4pt,
    breaklines=true,
    breakatwhitespace=true,
    columns=flexible,
    keepspaces=true,
    xleftmargin=1em,
    tabsize=2
]
Problem:
        {Problem Statement}

        IMPORTANT SOLUTION REQUIREMENTS:
        1. You MUST FIRST solve this problem using mathematical reasoning and symbolic calculations:
           - Use proper mathematical notation and symbols
           - Arrive at a final symbolic mathematical expression

        2. ONLY AFTER completing the mathematical solution:
           - Convert your final mathematical expression into Python code
           - The code must satisfy these requirements:
        {Specific Requirements for Python Code Answer}

        Code Format Requirements:
        1. Your solution MUST include the final executable Python code as required by the "Answer Requirements"
        2. You MUST wrap the final Python code between ```python and ``` tags
        3. Ensure the code is complete and can run independently
        4. The code should NOT contain ANY externally defined variables, including physical constants.
        5. The code MUST be concise and lightweight, faithfully and directly translating the final symbolic mathematical expression derived in step 1. Do NOT perform any further calculations or simplifications within the Python code itself.
        6. The code MUST NOT include any redundant elements such as conditional logic (if statements), exception handling (try/except blocks), or unnecessary checks.
\end{lstlisting}

\subsection{Prompt for Multi-Round Reasoning}
\label{app:multi-round prompt}
Initial Prompt
\begin{lstlisting}[
    basicstyle=\ttfamily\scriptsize,
    frame=single,
    framerule=0.4pt,
    breaklines=true,
    breakatwhitespace=true,
    columns=flexible,
    keepspaces=true,
    xleftmargin=1em,
    tabsize=2
]

[Problem Statement]

Provide detailed reasoning to solve this problem step by step.
\end{lstlisting}

Subsequent Iteration Prompt
\begin{lstlisting}[
    basicstyle=\ttfamily\scriptsize,
    frame=single,
    framerule=0.4pt,
    breaklines=true,
    breakatwhitespace=true,
    columns=flexible,
    keepspaces=true,
    xleftmargin=1em,
    tabsize=2
]
[General Contextual Instructions]

Original Problem:
[Problem Statement]

Summary from Previous Iteration's Reasoning:
[Previous Iteration's Conclusion]

Rethink the problem from here, then continue with detailed reasoning.
\end{lstlisting}

Summarization Prompt
\begin{lstlisting}[
    basicstyle=\ttfamily\scriptsize,
    frame=single,
    framerule=0.4pt,
    breaklines=true,
    breakatwhitespace=true,
    columns=flexible,
    keepspaces=true,
    xleftmargin=1em,
    tabsize=2
]

Conclude the reasoning process above and arrive at 1 final symbolic mathematical expression.

AFTER concluding the reasoning process:
   - Convert the final mathematical expression into Python code
   - The code must satisfy these requirements:
[Specific Requirements for Python Code Answer]

Code Format Requirements:
1. Your solution MUST include the final executable Python code as required by the "Answer Requirements"
2. You MUST wrap the final Python code between ```python and ``` tags
3. Ensure the code is complete and can run independently
4. The code should NOT contain ANY externally defined variables, including physical constants.
5. The code MUST be concise and lightweight, faithfully and directly translating the final symbolic mathematical expression derived.
6. The code MUST NOT include any redundant elements such as conditional logic, exception handling, or unnecessary checks.
\end{lstlisting}

\subsection{Prompt for Simple Weak Verifier}
\label{app:simple_verifier_prompt}
Verification Prompt (Randomly choose 1 of 3 below)

\begin{lstlisting}[
    basicstyle=\ttfamily\scriptsize,
    frame=single,
    framerule=0.4pt,
    breaklines=true,
    breakatwhitespace=true,
    columns=flexible,
    keepspaces=true,
    xleftmargin=1em,
    tabsize=2
]
1:      Question: {question}. Answer Requirements: {answer_requirements}.
        Below is a student's solution to the above problem. 
        Evaluate whether the student has correctly solved the problem and adheres to the answer requirements.
        Consider potential logical flaws, computational errors, or incorrect assumptions in the solution. 
        Solution: {detailed_solution}. 
        Please respond with a JSON object with only the following structure: 
        {{ "is_solution_correct": 'yes' or 'no' }}

2:      Question: {question}. Answer Requirements: {answer_requirements}.
        Here is a student's detailed solution to the problem.
        Assess the solution by checking if all necessary steps are followed and if any significant errors were made. 
        Your task is to determine whether the solution is valid and complete or if it contains major flaws.
        Solution: {detailed_solution}. 
        Provide your response strictly in the following JSON format:
        {{ "is_solution_correct": 'yes' or 'no' }}

3:      Question: {question}. Answer Requirements: {answer_requirements}.
        I include below a student solution to the above question and answer requirement. 
        Determine whether the student solution reaches the correct final answer in a correct fashion; 
        e.g., whether the solution makes two major errors that still coincidentally cancel out. 
        Please be careful and do not leap to conclusions without first reasoning them through. 
        Solution: {detailed_solution}. 
        Now summarize your response in a JSON format. Respond in the following format saying nothing else: 
        {{ "is_solution_correct": 'yes' or 'no' }}
\end{lstlisting}

Tie-Breaking Prompt

\begin{lstlisting}[
    basicstyle=\ttfamily\scriptsize,
    frame=single,
    framerule=0.4pt,
    breaklines=true,
    breakatwhitespace=true,
    columns=flexible,
    keepspaces=true,
    xleftmargin=1em,
    tabsize=2
]
  You are an expert physicist and AI assistant. Your task is to evaluate two attempted solutions to a given physics problem and determine which one is correct.

**Problem Statement:**
{problem_statement}

**Answer Requirement:**
{answer_requirement}

**Attempt 1:**
{attempt_1}

**Attempt 2:**
{attempt_2}

**Instructions:**

1.  **Analyze Problem and Requirements:** Carefully read the **Problem Statement** and the **Answer Requirement**. Understand what is being asked and what constraints or specific formats are required for the answer.
2.  **Evaluate Attempt 1:**
    * Assess the physical principles applied in Attempt 1. Are they appropriate for this problem?
    * Check the mathematical calculations and derivations. Are they correct?
    * Does the solution in Attempt 1 fully address the **Problem Statement**?
    * Does the final answer in Attempt 1 meet the **Answer Requirement** (e.g., units, significant figures, format)?
3.  **Evaluate Attempt 2:**
    * Assess the physical principles applied in Attempt 2. Are they appropriate for this problem?
    * Check the mathematical calculations and derivations. Are they correct?
    * Does the solution in Attempt 2 fully address the **Problem Statement**?
    * Does the final answer in Attempt 2 meet the **Answer Requirement** (e.g., units, significant figures, format)?
4.  **Determine Correct Solution:** Based on your evaluation, identify which attempt is correct.
5.  **Provide Justification:**
    * State clearly which attempt is correct (Attempt 1 or Attempt 2).
    * Provide a step-by-step explanation for why the chosen attempt is correct, referencing the physical principles, calculations, and adherence to the problem statement and answer requirements.
    * Explain why the other attempt is incorrect. Point out the specific errors in its application of principles, calculations, or its failure to meet the requirements. Be precise in your explanation of the flaws.

    Finally, Provide evaluation in compact JSON format with the following structure ONLY,
    remember to format your response using LaTeX notation for mathematical expressions, and follow the format for json
    
    {{"correct_attempt": "enter only number 1 or 2", "analysis": "Detailed explanation of the your justification"}}

    Ensure the output is *only* the JSON object, without any introductory text or markdown formatting around it.
    
\end{lstlisting}

\subsection{Prompt for SymPy-Augmented LLM Grading Agent}
\label{app:physics_grader_prompt}

The following prompt is used to instruct the LLM-based grading agent, which is a key component of our physics-augmented weak verifier pipeline as described in Stage~2 of Section~\ref{subsec:parallel_sampling_verifiers}. This agent is responsible for a step-by-step verification of candidate solutions, leveraging SymPy for validating mathematical calculations. 

\begin{lstlisting}[
    basicstyle=\ttfamily\scriptsize,
    frame=single,
    framerule=0.4pt,
    breaklines=true,
    breakatwhitespace=true,
    columns=flexible,
    keepspaces=true,
    xleftmargin=1em,
    tabsize=2
]
You are a grading assistant for theoretical physics problems. Your task is to grade a user-provided draft solution.

You need to carefully analyze every logical and computational step in the solution. During the analysis:
1.  Divide the entire mathematical derivation in the solution into sequentially numbered steps.
2.  Identify the key claims and formula usage within each step.
3.  For **each** numbered mathematical calculation step identified in step 1, you **must** construct a Python script containing SymPy code to perform that specific calculation. The script must print the final result to standard output.
4.  Use the available `run_sympy_script` tool to execute the Python (SymPy) script you constructed for each step.
5.  Carefully examine the Python script\'s output (STDOUT) and errors (STDERR) for **each step**. If there is an error message (STDERR), rewrite the script for that step and execute it again. If the result in STDOUT does not match the one in the solution for that step, this indicates a calculation error in that specific step.
6.  Perform the analysis and verification for all steps *before* grading. Based on your step-by-step analysis (including the correctness of the logical flow, accuracy of conceptual understanding) and the calculation results verified by the SymPy tool for each step, provide the final score.
7.  Ignore any final Python code within the solution; focus only on the preceding mathematical derivations and calculations identified in step 1.

Ensure your Python (SymPy) scripts are specific to the calculation, complete, directly runnable, and print the final verification result to standard output.
Before calling the tool for a specific step, briefly explain which step number and what calculation you intend to verify using SymPy.

A general note on string values within the JSON output:
- For literal backslashes (e.g., in LaTeX commands like `\\alpha` or `\\frac`), these must be escaped in the JSON string. So, `\\alpha` should be written as `\\\\alpha` in the JSON string value.
- For newline characters within a string (e.g., in multi-line script content), use `\\n`.
- For double quotes within a string, use `\\\"`.

Your final output **must** be a single JSON object. Do not add any text before or after the JSON object.
The JSON object should have the following structure:
{
  "sympy_verification": [
    // This is a list of objects, each corresponding to a mathematical calculation step that you verify using SymPy.
    // If no steps were verified using SymPy (e.g., if the solution has no calculable steps or all steps were skipped), this list can be empty.
    {
      "step_number": "integer: The sequence number of the calculation step you identified and verified.",
      "calculation_description": "string: A brief description of the calculation performed in this step.",
      "sympy_script_content": "string: The exact SymPy script you constructed and executed for this step.",
      "script_stdout": "string: The STDOUT captured from executing your SymPy script for this step.",
      "script_stderr": "string: The STDERR captured during execution (if any, otherwise empty string or null).",
      "is_correct": "boolean: true if the calculation in this step of the solution matches the SymPy output, false otherwise.",
      "error_explanation": "string: If \'is_correct\' is false, provide a detailed explanation of the error found in the solution step based on SymPy verification. If \'is_correct\' is true, this can be a brief confirmation or empty string."
    }
    // ... more objects for other verification steps
  ],
  "overall_score": "integer: Based on your comprehensive analysis, provide a numerical score of 0 or 1.",
  "general_feedback": "string: An overall assessment of the solution, highlighting its strengths, weaknesses, and the reasoning behind your scoring. This should summarize the conclusions drawn from your step-by-step analysis."
}

Example of a sympy_verification entry:
{
  "step_number": 1,
  "calculation_description": "Derivative of z with respect to eta.",
  "sympy_script_content": "import sympy\\n# ... rest of the script ...\\nprint(result)",
  "script_stdout": "z*(H + epsilon_prime/(2*epsilon))",
  "script_stderr": "",
  "is_correct": true,
  "error_explanation": "The calculation for z\\' is correct."
}
\end{lstlisting}

\section{Token count for sequential scaling method}
\label{app:token_sequential}

\begin{figure}[H]
    \centering
    \begin{minipage}{0.495\textwidth}
        \centering
        \includegraphics[width=\textwidth]{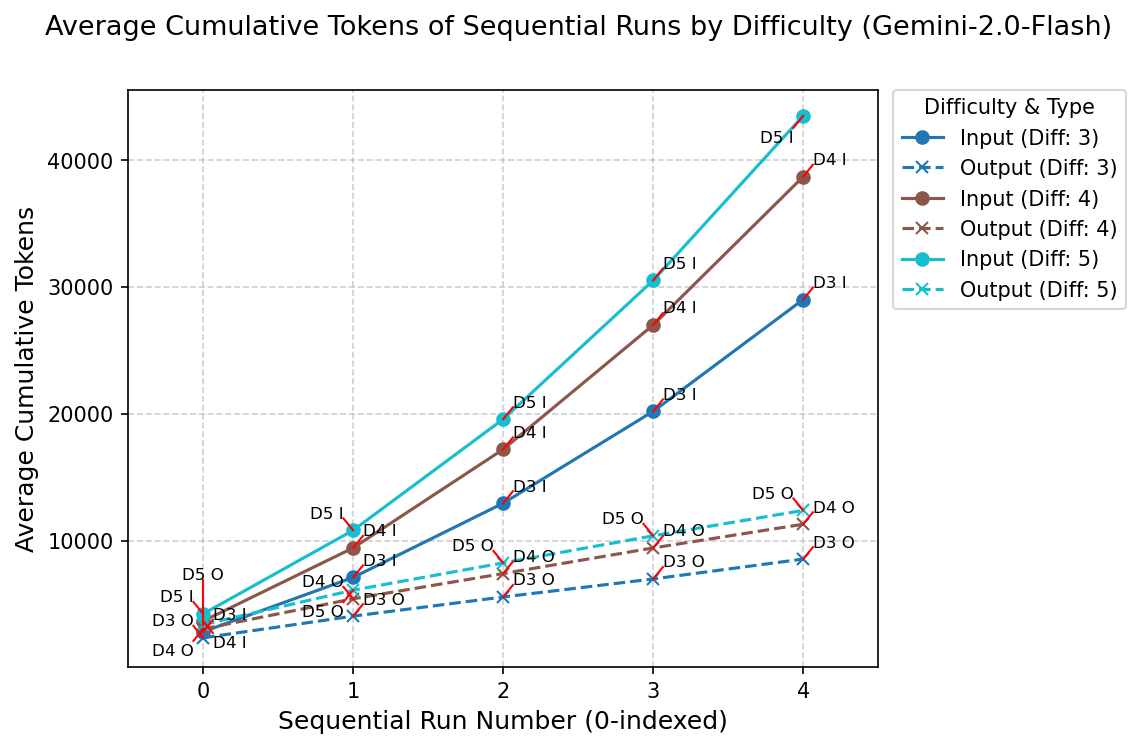}
    \end{minipage}
    \hfill
    \begin{minipage}{0.495\textwidth}
        \centering
        \includegraphics[width=\textwidth]{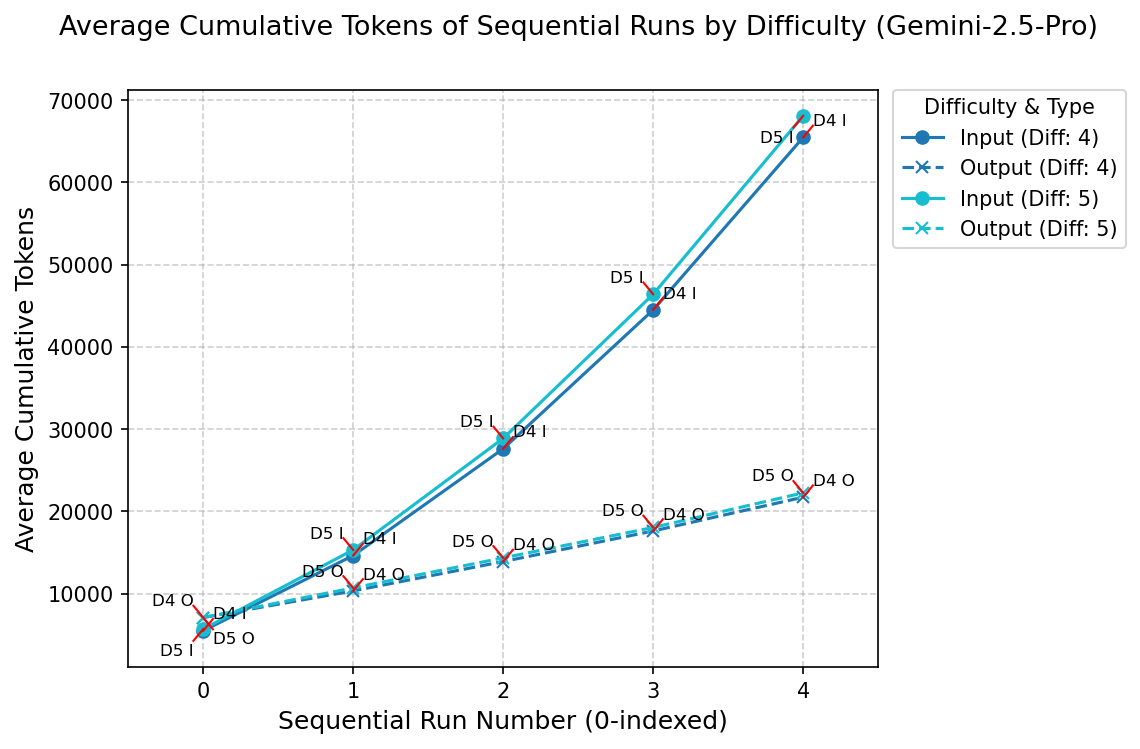}
    \end{minipage}
    \caption{Token Consumption Statistics by different models and difficulty levels, on the TPBench data set. More difficult problems require more tokens. Further, total token consumption grows approximately linearly for easy problems but somewhat more rapidly for difficult problems.}
\end{figure}

\section{Case Study: Stepwise SymPy-Enhanced Verification of the Halo Bias Problem}
\label{app:halo_bias_verification}

This appendix provides a detailed analysis of how our stepwise SymPy-augmented LLM verification framework systematically identified critical errors in the calculation of Gaussian moments within a candidate solution for a cosmological halo bias problem. This case study demonstrates the framework's capability to validate complex statistical mechanics derivations in physics by checking the steps with SymPy scripts.

\subsection{Problem Statement}

The problem requires deriving the effective bias of a clipped halo density field, a common procedure in modeling large-scale structure formation. It is one of the public level 5 problems of TPBench, and is often solved correctly by leading reasoning models.

\begin{mdframed}[style=solutionbox, frametitle={Halo Bias Problem Statement}]
In cosmology, large-scale cosmological dark-matter halo fields are biased tracers of the underlying Gaussian matter density $\delta_m$. We simulate a halo number density field by taking $n(\mathbf{x}) = \bar{n}\max(0,1+b\delta_m(\mathbf{x}))$, where the bare number density $\bar{n}$ and bare bias $b$ are specified constants, and $\delta_m$ is a Gaussian random field with zero mean and variance $\sigma^2$ in each pixel.

\textbf{Task}: Derive an equation to evaluate the effective bias $b_{\text{eff}}$ of the sampled halo field, defined by the relation $\langle\delta_h\rangle = b_{\text{eff}}\delta_L$ for a long-wavelength background perturbation $\delta_L$. The final expression should depend on the bare bias $b$ and the pixel variance $\sigma^2$.
\end{mdframed}

\subsection{Candidate Solution Overview}

The candidate solution attempts to derive the effective bias using the peak-background split formalism. This involves expressing the mean halo density and its response to a long-wavelength perturbation as expectation values over the Gaussian distribution of the matter density contrast $\delta_m$. The key steps in the candidate's derivation are outlined below.

\begin{mdframed}[style=solutionbox, frametitle={Candidate Solution Structure}]

The solution starts by expressing the mean halo density $\bar{n}_h$ as an expectation value over the Gaussian field $X \equiv \delta_m$. The clipping at zero means the integral is non-zero only for $1+bX > 0$, or $X > -1/b$. Let $\nu = -1/b$.
\begin{equation}
\bar{n}_h = \langle n(X) \rangle = \bar{n} \int_{-1/b}^{\infty} (1+bX) P(X) dX
\end{equation}
where $P(X)$ is a Gaussian PDF with mean 0 and variance $\sigma^2$. This is expanded into two integrals:
\begin{equation}
\bar{n}_h = \bar{n} \left[ \int_{\nu}^{\infty} P(X) dX + b \int_{\nu}^{\infty} X P(X) dX \right]
\end{equation}
The first term evaluates to $\int_{\nu}^{\infty} P(X) dX = \frac{1}{2} \text{erfc}\left(\frac{\nu}{\sigma\sqrt{2}}\right)$.

The solution then evaluates the first moment integral $I_1 = \int X P(X) dX$ over the truncated domain:
\begin{equation}
I_1 = \int_{-1/b}^{\infty} X P(X) dX = \frac{\sigma^2}{\sqrt{2\pi}} e^{-1/(2b^2\sigma^2)} \quad \textcolor{red}{\text{(ERROR HERE)}}
\end{equation}
Combining these gives the candidate's expression for the mean halo density:
\begin{equation}
\bar{n}_h = \bar{n} \left[ \frac{1}{2} \text{erfc}\left(-\frac{1}{b\sigma\sqrt{2}}\right) + b \frac{\sigma^2}{\sqrt{2\pi}} e^{-1/(2b^2\sigma^2)} \right]
\end{equation}

The effective bias is formulated using the peak-background split argument, which defines $b_{\text{eff}}$ as the response of the mean halo number density to a long-wavelength perturbation $\delta_L$:
\begin{equation}
b_{\text{eff}} = \frac{1}{\bar{n}_h} \frac{d \langle n | \delta_L \rangle}{d \delta_L} \Big|_{\delta_L=0}
\end{equation}
The derivative term is shown to be equivalent to:
\begin{equation}
\frac{d \langle n | \delta_L \rangle}{d \delta_L} \Big|_{\delta_L=0} = \int_{-1/b}^{\infty} \bar{n}(1+bX) \left( P(X)\frac{X}{\sigma^2} \right) dX = \frac{\bar{n}}{\sigma^2} \int_{-1/b}^{\infty} (X + bX^2) P(X) dX
\end{equation}

To evaluate the above, the solution requires the second moment integral $I_2 = \int X^2 P(X) dX$ over the truncated domain, for which it claims:
\begin{equation}
I_2 = \int_{-1/b}^{\infty} X^2 P(X) dX = \frac{-\sigma^2/b}{\sqrt{2\pi}} e^{-1/(2b^2\sigma^2)} + \sigma^3 \frac{1}{2} \text{erfc}\left(\frac{-1}{b\sigma\sqrt{2}}\right) \quad \textcolor{red}{\text{(ERROR HERE)}}
\end{equation}

Combining the (incorrect) intermediate moment calculations $I_1$ and $I_2$, the candidate assembles the numerator for the $b_{\text{eff}}$ expression:
\begin{equation}
D(\sigma, b) = \frac{\bar{n}}{\sigma^2} \left[ I_1 + b I_2 \right] = \bar{n} \left[ \frac{1+b\nu}{\sqrt{2\pi}} e^{-\nu^2/(2\sigma^2)} + \frac{b\sigma}{2} \text{erfc}\left(\frac{\nu}{\sigma\sqrt{2}}\right) \right]
\end{equation}
Substituting $\nu = -1/b$ correctly makes the term $(1+b\nu)$ equal to zero, which simplifies the expression to:
\begin{equation}
D(\sigma, b) = \bar{n} \frac{b\sigma}{2} \text{erfc}\left(-\frac{1}{b\sigma\sqrt{2}}\right)
\end{equation}
The effective bias is then $b_{\text{eff}} = D(\sigma, b) / \bar{n}_h$. Using the identity $\text{erfc}(-y) = 1 + \text{erf}(y)$ and simplifying, the candidate arrives at the final expression.

\textbf{Final Mathematical Expression}:
\begin{equation}
b_{\text{eff}}(\sigma, b) = \frac{ b\sigma \left(1 + \text{erfc}\left(\frac{1}{b\sigma\sqrt{2}}\right)\right) }{ \left(1 + \text{erfc}\left(\frac{1}{b\sigma\sqrt{2}}\right)\right) + \sqrt{\frac{2}{\pi}} b\sigma^2 \exp\left(-\frac{1}{2b^2\sigma^2}\right) }
\end{equation}

\end{mdframed}

\subsection{Step-by-Step Verification Process}

The following is a detailed verification of the mathematical derivations presented in the solution. Each step was computationally verified using the SymPy symbolic mathematics library in Python. The code used for each verification is provided.

\begin{enumerate}
    \item \textbf{Calculation of $\int_{\nu}^{\infty} P(X) dX$:}
    The solution evaluates the integral of the Gaussian probability density function $P(X) = \frac{1}{\sqrt{2\pi\sigma^2}} \exp\left(-\frac{X^2}{2\sigma^2}\right)$ from $\nu$ to $\infty$.
    \begin{itemize}
        \item \textbf{Expected Result:} $\frac{1}{2} \erfc\left(\frac{\nu}{\sigma\sqrt{2}}\right)$.
        \item \textbf{SymPy Verification:} The computation confirms the expected result.
        \item \textbf{Conclusion:} \textit{Correct}.
    \end{itemize}
    \begin{mdframed}[style=sympybox, frametitle={SymPy Verification of $\int_{\nu}^{\infty} P(X) dX$}]
        \begin{lstlisting}[style=mypython]
import sympy

X, nu = sympy.symbols('X nu', real=True)
sigma = sympy.Symbol('sigma', positive=True)

P_X = (1 / sympy.sqrt(2 * sympy.pi * sigma**2)) * sympy.exp(-X**2 / (2 * sigma**2))
integral_val = sympy.integrate(P_X, (X, nu, sympy.oo))

print(integral_val)
        \end{lstlisting}
    \end{mdframed}

    \item \textbf{Calculation of $I_1 = \int_{\nu}^{\infty} X P(X) dX$:}
    This step calculates the first moment of the truncated Gaussian distribution.
    \begin{itemize}
        \item \textbf{Stated Result:} $\frac{\sigma^2}{\sqrt{2\pi}} e^{-\nu^2/(2\sigma^2)}$.
        \item \textbf{SymPy Verification:} The computation yields $\frac{\sigma}{\sqrt{2\pi}} e^{-\nu^2/(2\sigma^2)}$.
        \item \textbf{Conclusion:} \textit{Incorrect}. The result stated in the solution contains an erroneous extra factor of $\sigma$.
    \end{itemize}
    \begin{mdframed}[style=sympybox, frametitle={SymPy Verification of $I_1 = \int_{\nu}^{\infty} X P(X) dX$}]
        \begin{lstlisting}[style=mypython]
import sympy

X, nu = sympy.symbols('X nu', real=True)
sigma = sympy.Symbol('sigma', positive=True)

P_X = (1 / sympy.sqrt(2 * sympy.pi * sigma**2)) * sympy.exp(-X**2 / (2 * sigma**2))
integrand = X * P_X
integral_val = sympy.integrate(integrand, (X, nu, sympy.oo))

print(integral_val)
        \end{lstlisting}
    \end{mdframed}

    \item \textbf{Internal Consistency of $\bar{n}_h$ Expression:}
    This step verifies the algebraic construction of $\bar{n}_h = \bar{n} \left[ \int_{-1/b}^{\infty} P(X) dX + b \int_{-1/b}^{\infty} X P(X) dX \right]$ using the solution's flawed intermediate results.
    \begin{itemize}
        \item \textbf{Stated Result:} $\bar{n} \left[ \frac{1}{2} \erfc\left(-\frac{1}{b\sigma\sqrt{2}}\right) + b \frac{\sigma^2}{\sqrt{2\pi}} e^{-1/(2b^2\sigma^2)} \right]$.
        \item \textbf{SymPy Verification:} The code correctly reconstructs the stated formula for $\bar{n}_h$ using the flawed value of $I_1$.
        \item \textbf{Conclusion:} \textit{Algebraically Consistent}.
    \end{itemize}
    \begin{mdframed}[style=sympybox, frametitle={SymPy Verification of $\bar{n}_h$ Construction}]
        \begin{lstlisting}[style=mypython]
import sympy

sigma_sym, b_sym, n_bar = sympy.symbols('sigma b n_bar', real=True)
sigma = sympy.Symbol('sigma', positive=True) 
b = sympy.Symbol('b', real=True) 
nu_val = -1/b

term1_integral_P_X = sympy.Rational(1,2) * sympy.erfc(-1 / (b * sigma * sympy.sqrt(2)))
term2_integral_X_P_X_solution = (sigma**2/sympy.sqrt(2*sympy.pi))*sympy.exp(-1/(2*b**2*sigma**2))

n_h_constructed = n_bar * (term1_integral_P_X + b * term2_integral_X_P_X_solution)

print(n_h_constructed)
        \end{lstlisting}
    \end{mdframed}

    \item \textbf{Calculation of the Derivative $\frac{d P(X|\delta_L)}{d \delta_L} \Big|_{\delta_L=0}$:}
    This step evaluates the derivative of a shifted Gaussian distribution.
    \begin{itemize}
        \item \textbf{Stated Result:} $P(X) \frac{X}{\sigma^2}$.
        \item \textbf{SymPy Verification:} The derivative is calculated as $\frac{X}{\sigma^3\sqrt{2\pi}} \exp\left(-\frac{X^2}{2\sigma^2}\right)$, which is equivalent to $P(X) \frac{X}{\sigma^2}$.
        \item \textbf{Conclusion:} \textit{Correct}.
    \end{itemize}
    \begin{mdframed}[style=sympybox, frametitle={SymPy Verification of $\frac{d P(X|\delta_L)}{d \delta_L} |_{\delta_L=0}$}]
        \begin{lstlisting}[style=mypython]
import sympy

X, delta_L = sympy.symbols('X delta_L', real=True)
sigma = sympy.Symbol('sigma', positive=True)

P_X_cond = (1/sympy.sqrt(2*sympy.pi*sigma**2))*sympy.exp(-(X - delta_L)**2/(2*sigma**2))
deriv_P_X_cond = sympy.diff(P_X_cond, delta_L)
deriv_at_0 = deriv_P_X_cond.subs(delta_L, 0)

print(deriv_at_0)
        \end{lstlisting}
    \end{mdframed}
    
    \item \textbf{Calculation of $I_2 = \int_{\nu}^{\infty} X^2 P(X) dX$:}
    This step calculates the second moment of the truncated Gaussian distribution.
    \begin{itemize}
        \item \textbf{Stated Result:} $\frac{\sigma^2 \nu}{\sqrt{2\pi}} e^{-\nu^2/(2\sigma^2)} + \frac{\sigma^3}{2} \erfc\left(\frac{\nu}{\sigma\sqrt{2}}\right)$.
        \item \textbf{SymPy Verification:} The computed integral is $\frac{\sigma \nu}{\sqrt{2\pi}} e^{-\nu^2/(2\sigma^2)} + \frac{\sigma^2}{2} \erfc\left(\frac{\nu}{\sigma\sqrt{2}}\right)$.
        \item \textbf{Conclusion:} \textit{Incorrect}. The solution has errors in the powers of $\sigma$ for both terms.
    \end{itemize}
    \begin{mdframed}[style=sympybox, frametitle={SymPy Verification of $I_2 = \int_{\nu}^{\infty} X^2 P(X) dX$}]
        \begin{lstlisting}[style=mypython]
import sympy

X, nu = sympy.symbols('X nu', real=True)
sigma = sympy.Symbol('sigma', positive=True)

P_X = (1 / sympy.sqrt(2 * sympy.pi * sigma**2)) * sympy.exp(-X**2 / (2 * sigma**2))
integrand = X**2 * P_X
integral_val = sympy.integrate(integrand, (X, nu, sympy.oo))

print(integral_val)
        \end{lstlisting}
    \end{mdframed}
    
    \item \textbf{Algebraic Simplification of $D(\sigma, b)$:}
    This step checks the algebraic simplification of $D(\sigma, b) = \frac{\bar{n}}{\sigma^2} [ I_1^{\text{sol}} + b I_2^{\text{sol}} ]$ using the solution's incorrect expressions for $I_1$ and $I_2$.
    \begin{itemize}
        \item \textbf{Target Result:} $\bar{n} \left[ \frac{1}{\sqrt{2\pi}} e^{-\nu^2/(2\sigma^2)} (1 + b\nu) + \frac{b\sigma}{2} \erfc\left(\frac{\nu}{\sigma\sqrt{2}}\right) \right]$.
        \item \textbf{SymPy Verification:} Algebraically combining the flawed $I_1$ and $I_2$ expressions correctly yields the target expression.
        \item \textbf{Conclusion:} \textit{Algebraically Consistent}.
    \end{itemize}
    \begin{mdframed}[style=sympybox, frametitle={SymPy Verification of $D(\sigma,b)$ Simplification}]
        \begin{lstlisting}[style=mypython]
import sympy

nu = sympy.Symbol('nu', real=True)
sigma = sympy.Symbol('sigma', positive=True)
b = sympy.Symbol('b', real=True)
n_bar = sympy.Symbol('n_bar', real=True)

I1_sol = (sigma**2 / sympy.sqrt(2*sympy.pi)) * sympy.exp(-nu**2 / (2*sigma**2))
I2_term1_sol = (sigma**2 * nu / sympy.sqrt(2*sympy.pi)) * sympy.exp(-nu**2 / (2*sigma**2))
I2_term2_sol = sigma**3 * sympy.Rational(1,2) * sympy.erfc(nu / (sigma * sympy.sqrt(2)))
I2_sol = I2_term1_sol + I2_term2_sol

D_constructed = (n_bar / sigma**2) * (I1_sol + b * I2_sol)
D_constructed_simplified = sympy.expand(sympy.simplify(D_constructed))

print(D_constructed_simplified)
        \end{lstlisting}
    \end{mdframed}
    
    \item \textbf{Substitution of $\nu = -1/b$ into $D(\sigma, b)$:}
    This step substitutes the lower integration limit $\nu = -1/b$ into the expression for $D(\sigma, b)$.
    \begin{itemize}
        \item \textbf{Expected Result:} $\bar{n} \frac{b\sigma}{2} \erfc\left(-\frac{1}{b\sigma\sqrt{2}}\right)$.
        \item \textbf{SymPy Verification:} Upon substituting $\nu = -1/b$, the expression simplifies exactly to the expected result.
        \item \textbf{Conclusion:} \textit{Correct}.
    \end{itemize}
    \begin{mdframed}[style=sympybox, frametitle={SymPy Verification of $D(\sigma,b)$ with $\nu=-1/b$}]
        \begin{lstlisting}[style=mypython]
import sympy

nu_sym = sympy.Symbol('nu', real=True)
sigma = sympy.Symbol('sigma', positive=True)
b = sympy.Symbol('b', real=True)
n_bar = sympy.Symbol('n_bar', real=True)

term_A = (1 + b*nu_sym)/sympy.sqrt(2*sympy.pi)*sympy.exp(-nu_sym**2/(2*sigma**2))
term_B = (b * sigma / 2) * sympy.erfc(nu_sym / (sigma * sympy.sqrt(2)))
D_prev_step_form = n_bar * (term_A + term_B)

D_substituted = D_prev_step_form.subs(nu_sym, -1/b)
D_substituted_simplified = sympy.simplify(D_substituted)

print(D_substituted_simplified)
        \end{lstlisting}
    \end{mdframed}

    \item \textbf{Final Simplification of $b_{\text{eff}}$:}
    This step verifies the final algebraic manipulation of the $b_{\text{eff}}$ expression.
    \begin{itemize}
        \item \textbf{Stated Result:} $\frac{ b\sigma (1 + \erfc(\frac{1}{b\sigma\sqrt{2}})) }{ (1 + \erfc(\frac{1}{b\sigma\sqrt{2}})) + \sqrt{\frac{2}{\pi}} b\sigma^2 \exp(-\frac{1}{2b^2\sigma^2}) }$.
        \item \textbf{SymPy Verification:} The sequence of substitutions and algebraic simplifications was confirmed to match the solution.
        \item \textbf{Conclusion:} \textit{Correct}. The algebraic simplification is valid, but it relies on mathematically incorrect intermediate results ($I_1, I_2$).
    \end{itemize}
    \begin{mdframed}[style=sympybox, frametitle={SymPy Verification of $b_{\text{eff}}$ Final Simplification}]
        \begin{lstlisting}[style=mypython]
import sympy

sigma = sympy.Symbol('sigma', positive=True)
b = sympy.Symbol('b', real=True)
y_var = sympy.Symbol('y', real=True)

# Starting expression for b_eff, substituting y = 1/(b*sigma*sqrt(2))
# and using erfc(-y) = 1 + erf(y)
num_erf = (b * sigma / 2) * (1 + sympy.erf(y_var))
denom_term1_erf = sympy.Rational(1,2) * (1 + sympy.erf(y_var))
denom_term2 = (b * sigma**2 / sympy.sqrt(2*sympy.pi)) * sympy.exp(-y_var**2)

# Multiply numerator and denominator by 2 and simplify
num_final = b * sigma * (1 + sympy.erf(y_var))
denom_final = (1 + sympy.erf(y_var)) + (sympy.sqrt(2/sympy.pi)*b*sigma**2)*sympy.exp(-y_var**2)

final_b_eff_expr_in_y = num_final / denom_final

# Substitute y back to get expression in b and sigma
y_definition = 1 / (b * sigma * sympy.sqrt(2))
final_b_eff_expr = final_b_eff_expr_in_y.subs(y_var, y_definition)

print(sympy.simplify(final_b_eff_expr))
        \end{lstlisting}
    \end{mdframed}
\end{enumerate}

\subsection{Complete Verification Output}

The agent's final verdict, based on a comprehensive check of all steps, is reproduced below.

\begin{mdframed}[style=agentfeedbackbox, frametitle={Agent's Final Verdict}]
\textbf{Overall Score}: 0/1

\textbf{General Feedback}: The solution attempts to derive the effective bias $b_{\text{eff}}$. The overall logical structure of the derivation (setting up integrals for mean density, using peak-background split for the derivative term, and combining them) is generally sound. The initial steps involving the zeroth moment and the PDF derivative were verified as correct.

\textbf{Key Error}: The primary errors lie in the calculation of the first and second moments of the truncated Gaussian distribution. Following the correct foundational steps, the candidate first makes an error in calculating $\int X P(X)dX$, which is incorrect by a factor of $\sigma$. This fundamental error is then compounded by similar errors in the second moment calculation. These errors propagate through the entire derivation, leading to an incorrect final expression for the effective bias $b_{\text{eff}}$.
\end{mdframed}

\section{Case Study: SymPy-Enhanced Verification of the One-Pole Problem}
\label{app:one_pole_verification}

This appendix provides a detailed analysis of how our SymPy-augmented LLM verification framework systematically identified a critical algebraic error in a candidate solution for the one-pole Bogoliubov coefficient calculation problem, a public level 5 TPBench problem that is generally not solved correctly even by the best models. While the symbolic verifier does identify a mistake, overall, its verification is less convincing than in the algebraically simpler case presented in App. \ref{app:halo_bias_verification}, indicating potential for future work on symbolic verification of mathematically difficult problems.

\subsection{Problem Statement}

The problem requires computing the Bogoliubov coefficient for a conformally coupled scalar field with the following specifications:

\begin{mdframed}[style=solutionbox, frametitle={One-Pole Problem Statement}]
Consider the conformally coupled scalar field $\phi$ with Lagrangian density:
\begin{equation}
\mathcal{L}=\frac{1}{2}\left[g^{\mu\nu}\partial_{\mu}\phi\partial_{\nu}\phi-\left(m^{2}-\frac{1}{6}R\right)\phi^{2}\right]
\end{equation}
in curved spacetime
\begin{equation}
ds^{2}=a^{2}(\eta)\left(d\eta^{2}-|d\vec{x}|^{2}\right)
\end{equation}
where the Ricci scalar is 
\begin{equation}
R=-6\frac{a''(\eta)}{a(\eta)}
\end{equation}
and $a$ satisfies the differential equation
\begin{equation}
\frac{d}{dt}\ln a=\Theta(t_{e}-t)H_{I}+\Theta(t-t_{e})\frac{H_{I}}{1+\frac{3}{2}H_{I}(t-t_{e})}
\label{eq:scalefactor_onepole_app}
\end{equation}
with $t_{e}$ a finite positive number, the $\Theta$ function having the steplike behavior
\begin{equation}
\Theta(t-t_{e})\equiv\begin{cases}
1 & t\geq t_{e}\\
0 & \text{otherwise}
\end{cases},
\end{equation}
and $t$ being the comoving proper time related to $\eta$ through
\begin{equation}
t=t_{e}+\int_{\eta_{e}}^{\eta}a(y)dy.
\end{equation}
The boundary condition for the differential equation is $a|_{t=t_{e}}=a_{e}$.

\textbf{Task}: In the limit that $k/(a_{e}H_{I})\rightarrow\infty$, using the steepest descent approximation starting from the dominant pole $\tilde{\eta}$ (with $\Re\tilde{\eta}>0$) of the integrand factor $\omega_{k}'(\eta)/(2\omega_{k}(\eta))$, compute the Bogoliubov coefficient magnitude $|\beta(k)|$ approximated as 
\begin{equation}
|\beta(k)|\approx\left|\int_{-\infty}^{\infty}d\eta\frac{\omega_{k}'(\eta)}{2\omega_{k}(\eta)}e^{-2i\int_{\eta_{e}}^{\eta}d\eta'\omega_{k}(\eta')}\right|
\label{eq:beta_original_app}
\end{equation}
for particle production, where the dispersion relationship is given by
\begin{equation}
\omega_{k}^{2}(\eta)=k^{2}+m^{2}a^{2}(\eta)
\end{equation}
with $0<m\lesssim H_{I}$. Use a one-pole approximation, which dominates in this limit.

\textbf{Answer Format}: Provide the answer as a Python function:
\begin{lstlisting}[style=mypython]
def abs_beta(k: float, a_e: float, m: float, H_I: float) -> float:
    pass
\end{lstlisting}
\end{mdframed}

\subsection{Candidate Solution Overview}

The candidate solution follows a systematic approach through the steepest descent method. We present here the key steps of the derivation that will be verified in detail. We note that the attempted solution is not very close to the correct expert solution which can be found in the TPBench paper. 

\begin{mdframed}[style=solutionbox, frametitle={Candidate Solution Structure}]
\textbf{Step 1: Mode Equation}\\
The solution begins by deriving the mode equation for $\chi_k(\eta)$ where $\phi = \chi/a$:
\begin{equation}
\chi_k''(\eta) + \omega_k^2(\eta) \chi_k(\eta) = 0
\end{equation}

\textbf{Step 2: Scale Factor Calculation}\\
Solving the differential equation \eqref{eq:scalefactor_onepole_app} yields:
\begin{equation}
a(\eta) = \begin{cases} 
\frac{a_e}{1 - a_e H_I \eta} & \eta < 0 \\ 
a_e \left( 1 + \frac{a_e H_I}{2} \eta \right)^2 & \eta \ge 0 
\end{cases}
\end{equation}

\textbf{Step 3-4: Integral Setup}\\
The Bogoliubov coefficient integral is expressed as $I = \int_{-\infty}^{\infty} d\eta f(\eta) e^{\phi(\eta)}$ where:
\begin{align}
f(\eta) &= \frac{\omega_k'(\eta)}{2\omega_k(\eta)}\\
\phi(\eta) &= -2i\int_{0}^{\eta} d\eta'\omega_{k}(\eta')
\end{align}

\textbf{Step 5: Pole Identification}\\
Poles occur when $\omega_k^2(\eta) = 0$, yielding:
\begin{itemize}
\item From $\eta < 0$ continuation: $\eta_p^{(1)} = \frac{1}{a_e H_I} \mp i \frac{m}{k H_I}$
\item From $\eta \ge 0$ continuation: Multiple poles with different imaginary parts
\end{itemize}

\textbf{Step 6: Dominant Pole Selection}\\
In the limit $k/(a_e H_I) \to \infty$, the dominant pole is identified as:
\begin{equation}
\tilde{\eta} = \frac{1}{a_e H_I} + i \frac{m}{k H_I}
\end{equation}

\textbf{Step 7-8: Residue and Approximation}\\
The residue at the pole is calculated as $\text{Res}[f,\tilde{\eta}] = 1/4$, leading to:
\begin{equation}
|\beta(k)| \approx \frac{\pi}{2} |e^{\phi(\tilde{\eta})}| = \frac{\pi}{2} \exp(2 \Im[J])
\end{equation}

\textbf{Step 9: Phase Integral (Critical Step)}\\
The candidate calculates $J = \int_0^{\tilde{\eta}} \omega_k d\eta'$ and claims:
\begin{equation}
\Im[J] = -\frac{1}{H_I} \sqrt{k^2 - (m a_e)^2} \quad \textcolor{red}{\text{(ERROR HERE)}}
\end{equation}

\textbf{Step 10: Final Result}\\
Using the above, the candidate concludes:
\begin{equation}
|\beta(k)| \approx \frac{\pi}{2} e^{-2k/H_I}
\end{equation}

\textbf{Python Implementation}:
\begin{lstlisting}[style=mypython]
import math

def abs_beta(k: float, a_e: float, m: float, H_I: float) -> float:
    """
    Computes |beta(k)| using steepest descent approximation
    in the limit k / (a_e * H_I) -> infinity.
    """
    result = (math.pi / 2.0) * math.exp(-2.0 * k / H_I)
    return result
\end{lstlisting}
\end{mdframed}

\subsection{Step-by-Step Verification Process}

\subsubsection{Steps 1--4: Foundation Verification}
The agent successfully verified the initial setup, including the mode equation derivation, scale factor calculation, and the formulation of the Bogoliubov coefficient integral. No errors were detected in these foundational steps.

\subsubsection{Step 5: Pole Location Calculation}
The agent verified the calculation of poles from the equation $\omega_k^2(\eta) = 0$. For the analytic continuation of $a(\eta) = a_e/(1 - a_e H_I \eta)$ (valid for $\eta < 0$), the poles are correctly identified as:

\begin{mdframed}[style=sympybox, frametitle={SymPy Verification of Pole Locations}]
\begin{lstlisting}[style=mypython]
# Solving (a_e / (1 - a_e * H_I * eta_p))^2 = -k^2 / m^2
# Result: eta_p = 1/(a_e*H_I) +/- I*m/(k*H_I)
\end{lstlisting}
\end{mdframed}


We note that this is an example where the agent did not write executable Python code but instead performed reasoning in the comments of the script. This sometimes happens and also allows us to spot mistakes.

\subsubsection{Step 6-8: Dominant Pole and Residue Analysis}
The verification confirmed:
\begin{itemize}
    \item The dominant pole in the specified limit is $\tilde{\eta} = \frac{1}{a_e H_I} + i\frac{m}{k H_I}$
    \item The residue calculation yielding $\text{Res}[f,\tilde{\eta}] = 1/4$ is correct
    \item The integral approximation framework is properly applied
\end{itemize}

\subsubsection{Step 9: Critical Error Detection}

The crucial error was discovered during the verification of the imaginary part of the phase integral:
\begin{equation}
    J = \int_0^{\tilde{\eta}} \omega_k(\eta') d\eta'
    \label{eq:phase_integral}
\end{equation}

The candidate solution claimed:
\begin{equation}
    \Im[J] = -\frac{1}{H_I} \sqrt{k^2 - (m a_e)^2}
    \label{eq:wrong_im_j}
\end{equation}

However, the SymPy verification revealed the correct result:

\begin{mdframed}[style=sympybox, frametitle={SymPy Script for $\Im[J]$ Verification}]
\begin{lstlisting}[style=mypython]
# Define symbols
a_e, H_I, k, m = symbols('a_e H_I k m', real=True, positive=True)
X = k / (m * a_e)  # Substitution variable, X > 1

# Based on the integral evaluation with sinh substitution:
# J = (m * a_e / H_I) * (sqrt(2) - (log(X) + I*sqrt(X**2 - 1)))
J_derivation = (m * a_e / H_I) * (sqrt(2) - (log(X) + I*sqrt(X**2 - 1)))

# Calculate the imaginary part
Im_J = im(J_derivation)
Im_J_simplified = simplify(Im_J.subs(X, k/(m*a_e)))

print(Im_J_simplified)
# Output: -sqrt(k**2 - a_e**2*m**2)/(H_I*a_e)
\end{lstlisting}
\end{mdframed}

The SymPy output clearly shows:
\begin{equation}
    \Im[J] = -\frac{1}{H_I a_e} \sqrt{k^2 - (m a_e)^2}
    \label{eq:correct_im_j}
\end{equation}

We note that the presented python script did not use Sympy for symbolic integration, but rather just checked whether a substitution was correctly performed (which it was not). Perhaps the LLM agent correctly recognized that the mistake was made in the substitution and demonstrated this with Python.

\subsection{Complete Verification Output}

For completeness, we reproduce the agent's final assessment:

\begin{mdframed}[style=agentfeedbackbox, frametitle={Agent's Final Verdict}]
\textbf{Overall Score}: 0/1

\textbf{General Feedback}: The solution presents a detailed derivation for the Bogoliubov coefficient $|\beta(k)|$ using the steepest descent method. The overall logical flow is mostly correct: identifying the mode equation, the scale factor, setting up the integral for $\beta(k)$, finding poles of the integrand, identifying the dominant pole, calculating the residue, and then approximating the integral.

\textbf{Key Error}: The primary error lies in the calculation of the imaginary part of the phase integral, $\Im[J]$ (Step 9). The solution's Step 9 states $\Im[J] = -\frac{1}{H_I}\sqrt{k^2 - (m a_e)^2}$, which is missing the factor $a_e$ in the denominator. This error propagates to the final expression for $|\beta(k)|$, resulting in an incorrect dependence on the physical parameters.
\end{mdframed}

\end{document}